# Full-Feature versus Limited-Input Machine Learning for Residential Energy Estimation: A Comparative Analysis of RECS and ResStock Under Realistic Input Constraints

Aditya Ramnarayan[1], Fatih Evren[2*], Patti Gunderson[2], Samuel Rosenberg[3]

[1] PhD Intern, Pacific Northwest National Laboratory

[2] Researcher, Building Systems Group at Pacific Northwest National Laboratory

[3] Data Scientist, Building Systems Group at Pacific Northwest National Laboratory

* Corresponding author, Fatih.evren@pnnl.gov

**Abstract:**

Residential energy estimates are frequently needed before detailed envelope characteristics, equipment efficiencies, infiltration measurements, sensor data, or long billing histories are available. Nevertheless, machine-learning models are commonly evaluated using information-rich feature sets that cannot be reproduced in practical deployment. This study quantifies the trade-off between predictive accuracy and input accessibility using two nationally representative U.S. residential-energy datasets: the survey-based Residential Energy Consumption Survey (RECS) and the simulation-based ResStock dataset. A structured machine-learning pipeline was developed for annual estimates of total energy, space heating, and space cooling. Full-feature models were first used to establish dataset-specific performance benchmarks. For total-energy estimation, the models were subsequently restricted to ten low-burden variables obtainable from occupants, administrative records, or location-based weather data without an on-site energy audit. Among the evaluated algorithms (CatBoost, XGBoost, LightGBM, Random Forest, and Neural Networks), CatBoost consistently achieved the highest predictive performance for the full-feature analysis, reaching $R^2 \approx 0.90$ for ResStock and $R^2 \approx 0.73$ for RECS. . When the feature set was restricted to ten homeowner-accessible inputs to simulate realistic deployment conditions, model performance converged to $R^2 \approx 0.61$ for RECS and $R^2 \approx 0.62$ for ResStock These findings demonstrate that increased algorithmic complexity cannot fully compensate for missing physical and behavioral information. However, within a more homogeneous ResStock cohort consisting of single-family detached, natural-gas-heated homes in Climate Zone 6A constructed between 2000 and 2010, a reduced-input model improved accuracy to $R^2 \approx 0.85$, demonstrating the value of targeted modeling for homogeneous populations. The results indicate that tree-based ensemble models can serve as high-fidelity emulators of national-scale residential energy datasets. However, careful consideration of feature availability, dataset origin (empirical vs. synthetic), and applicable use cases are also important. The findings recast residential-energy modeling as an accuracy–accessibility problem and support a tiered deployment strategy in which low-burden national models are used for screening and preliminary planning, cohort-specific models support targeted estimation, and detailed audits or calibrated simulations are retained for high-consequence building-level decisions.

# 1. Introduction

In 2024, the United States consumed approximately 94.2 quadrillion British thermal units (quads) of energy, representing a 1% increase relative to 2023 (U.S. Primary Energy Production, 2024 - U.S. Energy Information Administration (EIA)). Furthermore, the average price of electricity for the residential sector in the US was 16.48 cents/kWh, rising at a faster rate since 2020 than in the previous decade (U.S. Electricity Retail Price by Sector 2024). Globally, energy demand continues to accelerate in response to population growth, rapid urbanization, and the widespread adoption of modern technologies (Asif & Muneer, 2007). The built environment is crucial, as buildings contribute substantially to total energy consumption. Among these, residential buildings are one of the largest and fastest-growing sectors. As urbanization intensifies, accurately predicting heating and cooling demands in residential buildings has become essential not only for improving energy efficiency but also for minimizing utility costs and supporting grid stability (Behmane & Pakere, 2024). In the United States, the challenge of rising energy prices is particularly urgent, as it has a direct impact on investment decisions, household budgets, and economic stability for building owners and occupants. Consequently, it is of great practical and economic significance to develop fast, reliable, and scalable methods for forecasting residential energy consumption, encompassing total energy use, space heating, and space cooling.

Residential energy estimates are required for many purposes, including building design and large-scale regional planning. Existing methods for estimating building energy consumption primarily involve simulation and data-driven models (Amasyali & El-Gohary, 2018). Simulation tools such as EnergyPlus, Trace 3D Plus, and IES VE are grounded in thermodynamic principles and can effectively model building energy behavior from fundamental concepts. However, these tools require highly detailed inputs, including occupant behavior profiles, precise building geometry and material properties, HVAC equipment information, along with considerable computational resources (Higgins et al., 2024; Liu et al., 2022; Sun et al., 2022); often resulting in significant discrepancies between predicted and actual energy use, with studies indicating that actual consumption may be approximately 1.7 times higher than simulations suggest (Aste et al., 2022). These limitations have sparked interest in data-driven alternatives that can learn directly from observed patterns in building and energy data. Data-driven models leverage machine learning algorithms and historical energy and weather data to capture complex, nonlinear relationships between building characteristics and energy consumption without requiring detailed physical specifications. A model to predict U.S. household energy consumption using the 2015 RECS dataset was developed by (Cui et al., 2024) and further applied SHapley Additive exPlanations (SHAP) analysis to evaluate the influence of household features on energy usage. These models have proven superior due to their computational efficiency and enhanced accuracy, especially with the growing accessibility of building data (Wei et al., 2018; Zargarzadeh et al., 2024). Consequently, there has been a notable shift towards data-driven approaches in predicting energy consumption and exploring energy efficiency in buildings.

In data-driven methods, typically an extensive set of building properties is fed into the models. However, in many situations, estimating energy consumption is necessary when detailed information about a dwelling is unavailable. Housing agencies may need preliminary consumption estimates for units without complete billing histories; planners may need estimates for buildings not covered by energy-disclosure programs; and designers must compare alternatives before envelope and mechanical-system specifications are finalized. In these settings, the principal constraint is often not the availability of a sophisticated modeling algorithm, but the limited amount of reliable building information available when the estimate is required. Models that rely on readily obtainable inputs can therefore provide practical value for preliminary screening, portfolio planning, benchmarking, and prioritizing buildings for more detailed assessment (Kontokosta & Tull, 2017; Seo et al., 2022).

Prior studies have shown that reducing the number of model inputs does not necessarily lead to proportional losses in predictive accuracy. Feature-selection studies have identified diminishing returns from increasingly large feature sets, while other work has explicitly developed building-energy models under reduced or constrained information availability (Ding et al., 2021; Kontokosta & Tull, 2017; Parhizkar et al., 2021). However, these studies differ substantially in prediction target, spatial resolution, building population, and the way input availability is defined. This makes it difficult to infer whether the performance consequences of imposing the same practical information constraint would be similar across fundamentally different representations of a national residential building stock.

Feature reduction and limited-input modeling, however, address related but distinct problems. Conventional feature-selection methods begin with an information-rich dataset and identify a subset of variables that preserves or improves predictive performance. Such an approach can reduce model complexity, but it does not ensure that the retained variables are readily obtainable when the model is applied to a new dwelling. A deployment-oriented model instead begins with the availability of information as a constraint. The relevant inputs must be obtainable from occupants, administrative records, public data, or location-based information without requiring detailed audits or monitoring. Consequently, the practical question is not only which variables are most predictive, but how much predictive performance can be retained when the model is restricted to information that can realistically be collected at scale.

Despite this growing literature, an important question remains regarding whether practical input constraints have similar consequences across fundamentally different representations of a national residential building stock. Previous studies have demonstrated that useful predictive performance can be retained with reduced feature sets or under constrained data availability, but they differ substantially in prediction target, spatial scale, building population, data source, and the way accessible inputs are defined. Consequently, their reported performance changes are difficult to compare directly. Among the studies reviewed here, we did not identify a direct comparison in which the same predefined set of low-burden inputs is imposed on both a nationally representative empirical household dataset and a national physics-based building-stock simulation under a common machine-learning framework.

A related question concerns model performance within more homogeneous residential populations. National-scale models must represent substantial heterogeneity in climate, construction vintage, housing type, heating fuel, and other characteristics. Evaluating a restricted-input model within a more narrowly defined cohort can therefore provide insight into how predictive performance changes when part of this heterogeneity is controlled. Such cohort-specific performance is not directly equivalent to performance across the national population, but it can indicate whether more targeted modeling is useful for applications involving relatively homogeneous groups of dwellings.

**Purpose and Contribution**

To address these questions, this study develops a unified machine learning framework for forecasting residential energy consumption across three targets: total energy use, space heating, and space cooling. The analysis leverages two complementary, nationally representative datasets: the survey-based RECS dataset, which reflects real-world variability and measurement uncertainty, and the physics-based ResStock dataset, which provides internally consistent, high-resolution synthetic building simulations. By applying a consistent modeling pipeline across both datasets, this study enables a direct comparison of predictive performance and feature importance between empirical and simulated data sources.

The analysis first develops information-rich models using the detailed features available within each dataset for total energy use, space heating, and space cooling. The primary deployment-oriented analysis then focuses on annual total energy use and compares these full-feature benchmarks with models restricted to the same ten low-burden inputs in both datasets: current heating and cooling degree days, historical-average heating and cooling degree days, housing type, number of bedrooms, cooking fuel, main space-heating fuel, air-conditioning equipment, and water-heating fuel. They were selected because they represent climatic, structural, and equipment characteristics that can reasonably be supplied by occupants, obtained from administrative records, or derived from location without a detailed on-site energy audit. The ten-input experiment therefore does not seek the mathematically optimal or universally smallest feature subset. Instead, it evaluates the performance cost of imposing a predefined and practically motivated information constraint. The selected variables and resulting reduced-input performance are summarized in Tables 1,2 and 5.

The contributions of this work are threefold. First, it provides a systematic cross-dataset evaluation of machine learning models trained on empirical and synthetic residential energy data, highlighting differences in predictive performance and underlying drivers. Second, it directly quantifies the change in annual total-energy estimation performance when detailed, dataset-specific feature sets are replaced by the same limited set of ten practically obtainable inputs. Third, it evaluates reduced-input model performance within a more homogeneous ResStock cohort defined by climate zone, housing type, heating fuel, and construction vintage, providing insight into model performance for more targeted residential populations. Together, these analyses distinguish the predictive performance achievable under information-rich research conditions from that attainable under a more deployment-oriented input constraint.

The practical question addressed by the limited-input analysis is therefore not simply how accurately residential energy use can be estimated when detailed building information is available, but how much predictive capability remains when the information supplied to the model is restricted to variables that can reasonably be obtained without a detailed audit. This distinction is directly relevant to applications such as preliminary screening, portfolio-level assessment, and utility-allowance estimation, where models must often be applied to large numbers of dwellings for which detailed envelope, system, and operational characteristics are unavailable. A motivating application is the calculation of utility allowances for income-qualified households by Public Housing Authorities (PHAs); 3.4 million households in the U.S. received a total of $6.9 billion in 2017 alone (United States Government Accountability Office, 2024), where estimates must be derived from a small, standardized set of inputs and detailed envelope or system data are typically unavailable (Riley & Associates, 2013).

# 2. Literature Review

Machine-learning methods have been increasingly applied to building energy estimation because they can capture complex relationships among climatic conditions, building characteristics, equipment, and occupant-related variables. This section reviews machine-learning approaches to building-energy estimation, previous studies using the RECS and ResStock datasets, and research examining feature reduction and energy estimation under limited data availability. Particular attention is given to the distinction between reducing an existing information-rich feature set and developing models from variables selected according to their practical availability at deployment.

## 2.1. Machine Learning Techniques for Building Energy Prediction

There are two traditional methods for forecasting building energy use: physical (white-box) and data-driven (black-box) modeling (Amasyali & El-Gohary, 2022). Physics-based approaches rely on thermodynamic principles and are implemented through simulation platforms such as EnergyPlus, IDA ICE, ESP-r, IESVE, TRNSYS, and eQUEST (Lam et al., 2014). While these tools can produce detailed energy estimates, they are computationally expensive, require extensive building-specific input data, and are difficult to scale across large and heterogeneous building stocks (Foucquier et al., 2013). Data-driven methods utilize machine learning algorithms to analyze historical energy consumption, weather conditions, and occupancy patterns (Ramnarayan et al., 2025).

A diverse array of machine learning algorithms has been employed in building energy forecasting, as highlighted by reviews examining their strengths and limitations across various building types and climates ((L. Zhang et al., 2021); (Tien et al., 2022); (Gassar & Cha, 2020)). Linear regression methods, valued for interpretability, often fall short in predictive accuracy for residential heating and cooling loads due to the non-linear nature of energy consumption determinants, with studies noting higher mean squared errors compared to more complex methods ((Jhamb & Ahmed, 2023); (Z. Wang et al., 2020)). Tree-based ensemble methods, such as decision trees, random forests, and gradient boosting machines (GBMs), have gained traction for their ability to capture non-linear interactions while maintaining computational efficiency. Decision trees outperform linear regression in complex datasets (Mehdizadeh Khorrami et al., 2024), random forests exhibit robustness across varied datasets ((Moayedi et al., 2019); (Sajjad et al., 2020)), and GBMs consistently achieve low error rates while requiring less computational power than deep learning approaches ((Di Persio & Fraccarolo, 2023); (Sauer et al., 2022); (Asamoah & Shittu, 2025)). Meanwhile, neural networks, particularly convolutional neural networks (CNNs) and long short-term memory (LSTM) networks, excel in capturing temporal and spatial dependencies in energy data but entail high computational costs and demand large training sets, limiting their practicality in resource-constrained contexts ((Amasyali & El-Gohary, 2021); (Asamoah & Shittu, 2025)). Support vector machines (SVMs) have demonstrated competitive performance in smaller or high-dimensional datasets (Li et al., 2009) but tend to underperform compared to ensemble methods as dataset complexity increases (Gassar & Cha, 2020). The emergence of nationally representative building energy datasets, particularly the Residential Energy Consumption Survey (RECS) and ResStock, has catalyzed the training and validation of these methods, enhancing data-driven residential energy forecasting over the past decade.

### 2.2. Studies using the RECS Dataset

A recent study conducted by (Cui et al., 2024) analyzed data from the 2015 Residential Energy Consumption Survey (RECS) to identify key factors affecting Energy Use Intensity (EUI) in U.S. apartments and single-family homes. The research highlighted several significant features, including total square footage, the use of natural gas for space heating, prevailing weather conditions, and the age of the building, all of which play a crucial role in determining energy consumption patterns. (Goldstein et al., 2022) also analyzed the 2015 RECS dataset and identified housing quality, floor area, and home ownership status as the primary drivers of energy consumption. In a similar vein, (Korsavi et al., 2025) developed bottom-up data-driven models utilizing SHAP sensitivity analysis based on the 2020 RECS micro dataset to identify the key factors influencing energy consumption in U.S. residential buildings. Their research on single-family detached homes revealed that the most significant features positively correlated with national energy consumption include heating degree days (HDDs), energy-consuming space, the total number of rooms, the number of household members, and winter thermostat settings. (L. Wang et al., 2021) used artificial neural networks and the Monte Carlo method to analyze the 2015 RECS dataset. Their findings indicated a positive relationship between energy end-use and several factors, including total building area, number of rooms, number of windows, winter temperature with heating, level of insulation, and respondent's age. Conversely, they found a negative relationship between energy end-use and summer temperature with cooling. (Estiri & Zagheni, 2019) analyzed RECS data from 1987, 1990, 2005, and 2009 to study the relationship between age and energy consumption in the U.S. housing sector. They found that energy consumption tends to increase as the household head ages, even when accounting for other factors.

### 2.3. Studies using the ResStock Dataset

In their study, (Asamoah & Shittu, 2025) utilized the ResStock dataset to assess the performance of seven different machine learning models for predicting energy loads in residential buildings. The findings indicated that the gradient boosting model achieved the best balance between accuracy and computational efficiency under limited resource conditions. Meanwhile, neural networks emerged as the top-performing model, achieving R-squared values of 0.91 for cooling loads and 0.90 for heating loads, along with the lowest root mean square error (RMSE) scores of 7.06 for cooling and 16.86 for heating. (Lee & Drouin, 2025) utilized the ResStock dataset to introduce an innovative framework for high-resolution forecasting of residential heating and electricity demand through probabilistic deep learning models. Their models demonstrated a reduction in RMSE scores by 18.3% and 35.1% compared to those based solely on ResStock. By offering an open-source, scalable, and high-resolution platform for demand estimation and forecasting, this research enhances the tools available to grid planners, thereby contributing to the ability to improve efficiency of the U.S. building stock and achieving reduced cost objectives. (Bhavsar et al., 2023) successfully utilized the ResStock dataset to create an advanced data-driven surrogate model known as SBEM-E, which employs light gradient boosting to deliver highly accurate predictions of end-use energy loads in residential buildings. This innovative model achieved a remarkable 53% improvement in accuracy over a traditional decision-tree-based surrogate model and demonstrated an impressive nearly 10-fold reduction in computational time compared to conventional physics-based building energy models. (Landsman et al., 2024) utilized both the ResStock and ComStock datasets to effectively analyze regional building stocks. They provided detailed end-use load shapes and assessed lifecycle consumption, emissions, and costs.

However, the characteristics that make ResStock particularly useful for machine-learning development also distinguish it from empirical household data. Building characteristics and energy outcomes are generated within a physically consistent simulation framework, and the dataset contains many descriptors that may not be known for an actual dwelling without detailed inspection or building records. High predictive performance obtained from the complete ResStock feature space therefore represents an information-rich modeling condition and does not necessarily indicate that comparable performance can be achieved when only a small number of readily obtainable dwelling characteristics are available.

### 2.4. Feature Reduction and Limited-Input Energy Estimation

There are a limited number of studies that have investigated whether building energy models can maintain predictive performance while using fewer input variables. These studies generally arise from two related motivations: reducing statistical or computational complexity and reducing the amount of information that must be collected before an energy estimate can be produced. Although these objectives overlap, they represent different modeling problems.

Feature-selection and dimensionality-reduction studies typically begin with a relatively detailed dataset and identify the variables that contribute most strongly to predictive performance. Parhizkar et al. (2021), for example, applied principal component analysis to historical building-energy and meteorological data before developing several prediction models. Their results showed that removing less informative dimensions could decrease computational and storage requirements while maintaining or improving predictive performance (Parhizkar et al., 2021). Ding et al. (2021) examined 43 building-related features using recursive feature elimination and found that the cumulative contribution of the ten most influential features exceeded 80% for all six machine-learning algorithms evaluated. They also observed diminishing marginal improvements as additional features were added and found that tree-based algorithms could attain satisfactory performance using fewer predictors (Ding et al., 2021).

Kontokosta and Tull (2017) reported a similar pattern at the urban scale. Using energy-disclosure records together with property and zoning data for New York City, they sequentially evaluated feature sets for estimating building energy consumption. Most of the reduction in prediction error was associated with the first few selected variables, and model accuracy generally ceased to improve after approximately six features. For city-wide predictions, adding additional variables could even decrease performance because relationships learned from the disclosure sample did not necessarily generalize to the broader building population (Kontokosta & Tull, 2017). These studies demonstrate that larger feature sets do not necessarily produce better models.

Seo et al. (2022) addressed the issue directly for residential buildings. Their study was motivated by the difficulty of obtaining drawings and detailed envelope information for existing homes. Input combinations were progressively simplified to eliminate measurements of doors, glazing, roof and floor area, wall and window area, and ceiling height. The resulting approach could be applied using only heating area and construction year from preliminary building records, with thermal characteristics inferred from construction year, thereby eliminating the need for a detailed field survey. The resulting model was applied to old houses occupied by low-income households in South Korea and demonstrated that useful heating-demand estimates could be obtained with substantially lower information requirements (Seo et al., 2022).

Lin et al. (2022) examined annual residential electricity and natural-gas consumption using data collected from houses in Oshawa, Canada. The study evaluated several feature-set sizes, and their results showed that useful predictive performance could be retained after substantial reductions in the number of variables, although performance depended on both the selected algorithm and feature set. The analysis was based on 83 houses for electricity and 73 for natural gas after incomplete observations were removed, and the authors noted that additional data would be required to extend the models beyond the study region (Lin et al., 2022).

Hsu (2014) examined the value of different types of building information by comparing models using combined benchmarking and engineering-audit information, benchmarking information alone, and audit information alone. The full-information and benchmarking-only models explained nearly identical proportions of the variation in energy use ($R^2 = 0.8271$ and 0.8268, respectively), whereas the audit-only model achieved $R^2 = 0.5270$. The results demonstrate that increasing the technical detail of the available information does not necessarily improve predictive performance and that the type of information available can be as important as its quantity (Hsu, 2014).

Recent work has extended limited-data approaches to considerably larger residential stocks. Sheng et al. (2025) developed an automated residential energy assessment method for approximately 143,000 properties in Sheffield. Using spatial, morphological, and thermal characteristics, the model achieved an $R^2$ of 0.828, with floor area and external-wall characteristics among the dominant predictors. The authors also noted, however, that reliance on energy-performance-certificate data introduced potential biases and that additional building and behavioral information could improve robustness (Sheng et al., 2025).

Colverd et al. (2025) more explicitly examined energy-estimation performance under different levels of information availability. Using data for approximately 609,000 postcodes in England and Wales, the authors developed a sequence of models ranging from information-rich configurations to models based on domain-invariant and globally scalable variables. Their full models achieved $R^2$ values of 0.93 and 0.89 for annual gas and electricity consumption, respectively. A model based on a substantially reduced set of domain-invariant inputs retained an $R^2$ of 0.87 for gas despite an 80% reduction in variables. Under more restrictive data availability, performance decreased, while the addition of urban-form and socioeconomic information progressively recovered part of the lost accuracy. This work provides particularly direct evidence that predictive performance depends on the information available at deployment, although its unit of analysis was small-neighborhood postcode aggregates rather than individual households.

Taken together, this literature demonstrates that useful building-energy estimates can often be obtained with substantially fewer variables than are available in information-rich research datasets. However, the consequences of restricting input information vary with prediction target, building population, spatial resolution, algorithm, and underlying data source. Feature reduction after a comprehensive dataset has been assembled is also conceptually different from defining model inputs in advance according to what can realistically be obtained for a new dwelling. Among the studies reviewed here, we did not identify a direct comparison in which the same predefined set of low-burden inputs is evaluated across both nationally representative empirical residential data and physics-based national building-stock simulations under a common modeling framework.

Building on this literature, the present study applies a consistent machine-learning pipeline to RECS and ResStock to estimate annual total energy use, space heating, and space cooling under information-rich conditions. The analysis then focuses on annual total energy use to compare these full-feature benchmarks with models restricted to the same ten low-burden inputs across both datasets, thereby quantifying the change in predictive performance associated with practical information constraints. Finally, a cohort-specific analysis evaluates reduced-input performance within a more homogeneous ResStock subset defined by climate zone, housing type, heating fuel, and construction vintage. Together, these analyses provide a comparative assessment of residential energy estimation under information-rich and deployment-constrained conditions.

## 3. Methodology

This study employs two nationally representative residential energy datasets, the 2020 Residential Energy Consumption Survey (RECS) (U.S. Energy Information Administration - EIA - Independent Statistics and Analysis, n.d.-a) and 2024.2 ResStock (ResStock - NLR, n.d.), to train, validate, and test machine learning models capable of emulating observed consumption patterns and forecasting energy use for previously unseen households. By developing this modeling framework, this study compares the datasets and characterizes their similarities and differences. Following validation and cohort-specific calibration, the resulting data-driven methodology can be applied to forecast residential energy consumption in contexts where empirical or simulated data are unavailable.

An important application of this work is the calculation of utility allowances for income-qualified households, which frequently face service disconnections. To mitigate these disruptions, Public Housing Authorities (PHAs) must accurately account for expected energy use when determining utility allowances. In many programmatic settings, these estimates must be derived using a limited and standardized set of inputs, typically restricted to basic building characteristics and location-dependent variables. Detailed information on envelope properties, system efficiencies, or occupancy behavior is often unavailable. This constraint motivates the reduced-feature modeling scenarios examined in this study, which are designed to reflect practical deployment conditions rather than idealized data availability. The modeling approach developed here can be used to support estimations that PHAs rely on to establish appropriate allowances for subsidized housing units. The methodology section details the analytical procedures, machine learning techniques, and case-study implementations used throughout this work.

Largely similar analyses were carried out on both the RECS and ResStock datasets with the idea of using these datasets to develop benchmark machine learning models for energy performance forecasting, as a reasonably accurate alternative energy use prediction method that is simpler, faster, and less expensive than gathering empirical data or developing bespoke energy models. The methodology encompasses a comprehensive framework that integrates data preprocessing, feature engineering, manual feature selection, hyperparameter finetuning and optimization, and explainability analysis. This approach aims to develop a robust, interpretable forecasting pipeline, thereby enhancing the reliability and transparency of predictive outcomes. Sections 3.1 and 3.2 discuss all the analyses carried out on the RECS and ResStock datasets, respectively. To further assess the generalizability of the proposed models under realistic data constraints, a reduced-feature analysis was conducted using only 10 attributes typically accessible to homeowners. Additionally, the robustness of the models was examined by slicing the data by key residential characteristics, including climate zone, housing type, heating fuel type, and building age.

### 3.1. Residential Energy Consumption Survey (RECS)

In 2020, over half (52%) of an average household's annual energy consumption was dedicated to just two uses: space heating and air conditioning (*Use of Energy in Homes - U.S. Energy Information Administration (EIA)*). These energy needs are primarily seasonal, are highly energy-intensive, and can vary based on factors such as geographic location, the size and structure of the home, and the types of equipment and fuels used. Figure 1 shows the breakdown of end-use energy consumption by US households.

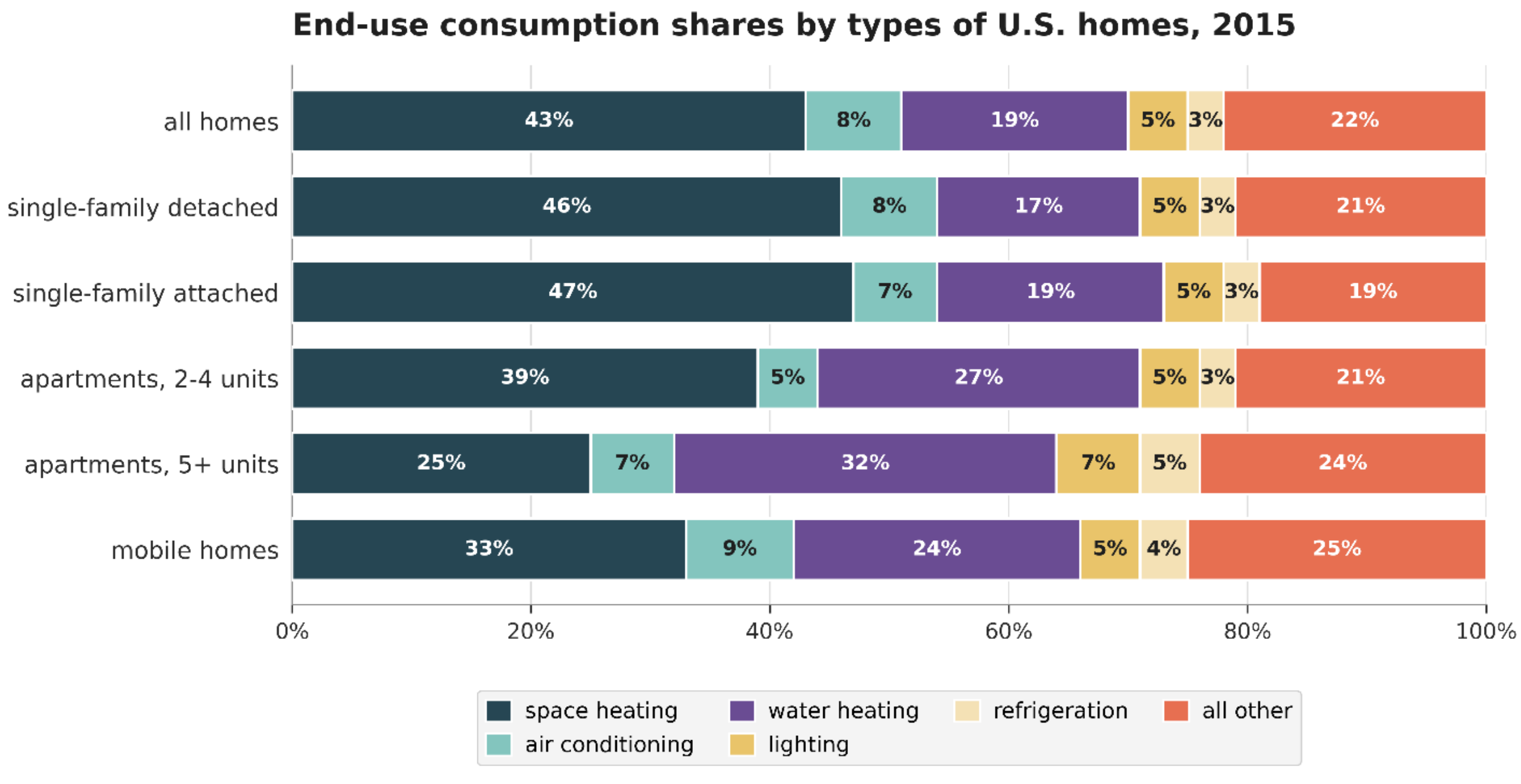

Figure 1: End-use energy consumption by US households (Use of Energy in Homes - U.S. Energy Information Administration (EIA))

The RECS dataset is particularly well-suited for research in this domain, as it offers comprehensive and detailed insights into household characteristics and energy consumption patterns (U.S. Energy Information Administration - EIA - Independent Statistics and Analysis, n.d.-b). It is considered a nationally representative cohort with comprehensive geographic coverage across the United States. It encompasses four distinct census regions, namely the Northeast, Midwest, South, and West, and further delineates these regions into ten specific census divisions: New England, Middle Atlantic, East North Central, West North Central, South Atlantic, East South Central, West South Central, Mountain North, Mountain South, and Pacific. This extensive classification ensures a thorough representation of the variety of characteristics present across the nation. This research leverages the 2020 RECS microdata provided by the U.S. Energy Information Administration (EIA). Based on responses from 18,496 households, this dataset is designed to reflect the energy profiles of approximately 123.5 million individual homes and their occupants (U.S. EIA, 2023c). It encompasses over 300 attributes covering various topics, including household characteristics, appliance usage, electronic devices, heating and cooling systems, lighting, and demographic information (U.S. EIA, 2023b).

#### 3.1.1. Feature Selection

The raw RECS dataset consists of 800 features and ~18,500 data points, making the dataset highly dimensional and reducing overfitting by retaining the most influential predictors. In previous studies (Estiri & Zagheni, 2019; L. Wang et al., 2021; Yun & Steemers, 2011), an initial set of 23 features most closely aligned with energy use was selected based on researchers' professional experience and specific objectives for energy consumption forecasts. A widely used method in machine learning for assessing correlations among variables is the Pearson correlation coefficient. This statistical measure effectively quantifies both the strength and the direction of a linear relationship between two variables (Benesty et al., 2008). A key pitfall of relying solely on this metric is its inability to capture non-linear relationships. In the context of the present study, the Pearson correlation methodology was initially employed; however, the top 15 features identified using this method as having the highest correlation with the target variable lacked intuitive significance. Consequently, this approach was deemed inadequate and subsequently discarded as a methodological strategy. Another study involving the 2020 RECS dataset, (Cui et al., 2024, 2025) used the Spearman correlation approach to determine the most ideal features and drivers of energy consumption prediction. The Spearman correlation coefficient evaluates the strength and direction of monotonic relationships between two variables, offering greater flexibility than the Pearson correlation, which is confined to linear relationships. The study (Cui et al., 2024) addressed the database's high dimensionality by combining Spearman correlation and p-value approaches to filter for variables exhibiting multicollinearity. To classify Energy Use Intensity (EUI), they retained household features with a correlation coefficient (R) of 0.10 or higher. Among pairs of features with an R of 0.40 or greater, the feature with the lower correlation to EUI was excluded to mitigate multicollinearity. Additionally, they applied a significance threshold of $p < 0.05$ to evaluate the relationship between household features and EUI. This rigorous filtering reduced the dataset to 14,070 samples and 25 relevant features, eliminating 605 low-correlation household attributes. However, the Spearman correlation's reliance on ranks may limit its ability to detect complex, non-monotonic relationships, which are critical for feature selection. Consequently, in this study, the features that showed up post-filtering (using the approach by (Cui et al., 2024, 2025)) were not sufficient to account as factors and drivers for use as the base for forecasts and hence were discarded.

In their study, (Korsavi et al., 2025) used an iterative approach in feature selection for energy forecasting. Using expert judgement, they included additional features over and above the ones used by (Estiri & Zagheni, 2019; L. Wang et al., 2021; Yun & Steemers, 2011) that could potentially affect residential building energy consumption, leading to an aggregate of 75 features in their first attempt. In their study, different machine learning models were trained and evaluated utilizing all 75 features, followed by SHAP sensitivity analysis to identify key influencers. Features deemed to have negligible relationships with energy use, based on SHAP values, were eliminated, yielding a final set of 41 features that maximized predictive performance ($R^2$) while minimizing error metrics (RMSE, MSE, and MAE). For comparison, the predictive performance achieved by (Korsavi et al., 2025) was the highest the authors found in literature so far, confirming this as a reasonable baseline approach for predicting energy performance using the RECS dataset. Building on previous research, this study applied a unique set of 20 selected features to the list, constituting 61 features in total as seen in Table 1, and combined them with a feature engineering approach (discussed further in Section 3.1.3) and then employed a Random Forest Regressor to select the 50 features with the highest correlation to energy use to ensure that the model is robust and generalizable.

*Table 1: List of features used for the ML model development for the RECS Dataset*

| | | | | |
|---|---|---|---|---|
| HDD65 | CDD65 | TYPEHUQ | YEARMADERANGE | BEDROOMS |
| DOOR1SUM | WINDOWS | TYPEGLASS | ORIGWIN | WINFRAME |
| ADQINSUL | DRAFTY | EQUIPM | FUELHEAT | AIRCOND |
| TYPETHERM | FUELH2O | NHSLDMEM | LGTINCAN | TOTSQFT_EN |
| DBT1 | DBT99 | IECC_climate_code | HDD30YR_PUB | CDD30YR_PUB |
| CELLAR | CRAWL | CONCRETE | BASEOTH | BASEFIN |
| ATTIC | ATTICFIN | TYPERFR1 | RANGEFUEL | TELLWORK |
| EMPLOYHH | DISHWASH | DRYRFUEL | OTHROOMS | TOTROOMS |
| TREESHAD | SIZRFRI1 | AGERFRI1 | ICE | FREEZER |
| AGEFRZR | OVENFUEL | TOPFRONT | HUMIDTYPE | NUMWWAC |
| DEHUMTYPE | SQFTEST | SQFTRANGE | NUMDLHP | ELPAY |

| NGPAY | ATHOME | MONEYPY | STORIES | ROOFTYPE |
|---|---|---|---|---|
| EQUIPAGE | | | | |

### 3.1.2. Data pre-processing

The data preprocessing steps include outlier detection and elimination, one-hot encoding for categorical features, and constraining numerical features. A one-hot encoding method was utilized to transform each categorical feature into multiple binary numerical features, ensuring a distinct representation for each category. In this encoding scheme, each category is represented as a vector, where a value of 1 ("hot") indicates the presence of the category, and a value of 0 ("cold") signifies its absence. One-hot encoding was specifically selected to improve interpretability and to effectively capture the directional influence of categorical variables on energy consumption (Okada et al., 2019). Following this, feature scaling was applied to all values by subtracting the mean and dividing by the standard deviation (z-score normalization) to ensure uniform scaling and compatibility with various modeling techniques. This process involved rescaling the numerical features to have a mean of zero and a standard deviation of one (Fromer, 2019). To address outlier removal, this study employed a K-means (k=5) clustering method to group data points into k distinct categories based on their similarities. This ensured that only data points closely aligned with each group's primary patterns were retained for analysis (Jain, 2010).

### 3.1.3. Feature Engineering

Feature engineering is a crucial part of developing machine learning models, as it acts as a vital link between raw data and a model's ability to make predictions (Reid Turner et al., 1999). For this study, using the set of 61 features selected for the analysis as described in Section 3.1.1, five additional features were engineered, whose formulae are as follows:

- Heating Load = Heating Degree Days (HDD) x Square Footage
- Cooling Load = Cooling Degree Days (CDD) x Square Footage
- Home Age = Year of consideration – average of year range in which constructed
- Square footage per person = Total square footage/ number of household members
- Draftiness score = 5 – response for variable 'DRAFTY' in dataset

### 3.1.4. Machine Learning Models

This study employs Ensemble tree-based machine learning algorithms, i.e., CatBoost, XGBoost, LightGBM, and Random Forest, owing to their effectiveness in capturing complex nonlinear relationships and managing categorical and missing data. Ensemble tree-based methods, such as Random Forests and Gradient Boosted Decision Trees (GBDTs), improve predictive performance by aggregating the outputs of multiple weak learners to reduce variance and bias (Breiman, 2001). By leveraging techniques like bagging and boosting, these models achieve state-of-the-art accuracy and robust generalization on heterogeneous tabular datasets. These ensemble learning methods are especially beneficial for structured datasets, like those used in energy consumption analysis. To complement these architectures, Lasso Regression was also implemented, which is a linear approach that uses L1 regularization to penalize the absolute magnitudes of coefficients, effectively performing automated feature selection by shrinking less influential variables to zero (Tibshirani, 1996). This ensures the model remains conservative and reduces the risk of overfitting in high-dimensional spaces. Random Forest enhances accuracy and robustness by using bagging (bootstrap aggregation), which combines the predictions of multiple decision trees (Kontokosta & Tull, 2017). CatBoost is distinguished by its capability to manage categorical features natively, eliminating the need for extensive hyperparameter tuning. This characteristic enhances its usability and effectiveness in various applications (Pan & Zhang, 2020). It effectively manages multicollinearity using an ordered boosting algorithm and built-in regularization techniques, which help mitigate overfitting and lessen the impact of highly correlated features on model predictions (Hancock & Khoshgoftaar, 2020). LightGBM and XGBoost use boosting techniques to enhance predictive performance by combining weak learners and correcting previous errors (D. Zhang & Gong, 2020). Ultimately, these ensemble architectures provide the necessary flexibility and predictive power to navigate the high dimensionality and non-linear dependencies inherent in large-scale energy consumption datasets.

**Performance Metrics**

Model evaluation is a critical step in the machine learning workflow, as it helps assess both the predictive capability and robustness of the developed model. To thoroughly analyze the proposed model's performance, the data was split into two sets: 80% for training and 20% for testing. Several commonly used regression metrics were employed, including R-squared ($R^2$), Mean Absolute Error (MAE), and Root Mean Squared Error (RMSE). This section presents the mathematical formulations of these metrics and discusses their significance in interpreting the model's predictive accuracy (*Hands-On Machine Learning with Scikit-Learn, Keras, and TensorFlow, 2nd Edition*, n.d.).

R-squared ($R^2$)

The coefficient of determination, known as R-squared, is a statistical measure that quantifies the proportion of variance in the dependent variable that can be predicted from the independent variables. R-squared serves as an indicator of the "goodness of fit" and is defined as:

$$R^2 = 1 - \frac{\sum_{i=1}^{n}(y_i - \hat{y}_i)^2}{\sum_{i=1}^{n}(y_i - \bar{y})^2} \tag{1}$$

where $y_i$ is the actual value, $\hat{y}_i$ is the predicted value, and $\bar{y}$ is the mean of the actual values. An $R^2$ value close to 1 indicates that the model effectively explains the variance in the target variable, suggesting a good fit. However, it's important to remember that $R^2$ can be misleading; it doesn't detect systematic bias in predictions (e.g., consistent underestimation) and can artificially increase with the addition of irrelevant predictors.

Mean Absolute Error (MAE)

The Mean Absolute Error (MAE) quantifies the average magnitude of errors in a set of predictions, disregarding their direction. This linear score treats all individual differences equally. Its resilience to outliers makes MAE a valuable metric, particularly when it is important not to disproportionately penalize large errors. The calculation of MAE is as follows:

$$MAE = \frac{1}{n}\sum_{i-1}^{n}|y_i - \hat{y}_i| \tag{2}$$

The result of MAE is in the same unit as the original target variable, which makes it straightforward and interpretable for stakeholders. A lower MAE indicates better model performance.

Root Mean Square Error (RMSE)

The Root Mean Squared Error (RMSE) represents the square root of the average of the squared differences between predicted and actual values. By squaring the errors, RMSE imposes a greater penalty on larger, more significant errors, making it particularly sensitive to outliers. The formula for calculating RMSE is as follows:

$$RMSE = \sqrt{\frac{1}{n}\sum_{i-1}^{n}(y_i - \hat{y}_i)^2} \tag{3}$$

Similar to Mean Absolute Error (MAE), Root Mean Square Error (RMSE) is expressed in the same units as the predicted variable. This allows for an intuitive interpretation and easy comparison across different models. Generally, a model with lower RMSE is considered to have better predictive accuracy, especially when large errors are not acceptable.

### 3.1.5. SHAP analysis

Interpreting machine learning models presents a significant challenge in the realm of energy prediction, as these models frequently operate as black boxes with limited transparency. This lack of interpretability not only poses difficulties for researchers seeking to understand the factors driving the predictions but also obstructs adoption by stakeholders, such as building owners, who need actionable insights rather than abstract outputs (Tian, 2024). Addressing this limitation necessitates the development of methods that can clarify the model's behavior and identify the influence of input features on its predictions (Panigrahi et al., 2025). SHAP (SHapley Additive exPlanations) is a method that explains the output of any machine learning model. It calculates the contribution of each feature to a single prediction, offering clear and interpretable insights (Ponce-Bobadilla et al., 2024). For the SHAP Analysis, features are ranked from most important (top) to least important (bottom). The color of each dot (Red/Pink = High, Blue = Low) tells if the value of the feature is low or high for that household. This study utilized SHAP to identify and interpret the key features driving predictions for total energy consumption, including space heating and cooling, to ensure transparency and enhance understanding of the model's decision-making process.

### 3.2. ResStock

The ResStock dataset is a comprehensive physics simulation model representing the residential building inventory in the United States. It leverages housing characteristics and geographical information from multiple sources to create a network of probability distributions that inform inputs for building energy models. The dataset includes simulations of approximately 550,000 building energy models, providing a statistical representation of residential structures in the contiguous United States. Additionally, the data integrates around 900 weather stations to illustrate the weather conditions that affect building energy consumption (*ResStock - NLR*).

#### 3.2.1. Feature selection and weather data addition

Similar to the RECS dataset, the ResStock dataset is highly dimensional, with plenty of features, making the selection of the most appropriate and accurate features a crucial step in the energy performance forecasting pipeline (Zhao & Magoulès, 2012). The ResStock dataset lacked weather-related features such as meteorological data in the form of heating and cooling degree days (HDD and CDD), similar to the RECS dataset. To maintain consistency across the two datasets, it was desired to use the HDD and CDD for the machine learning analysis. For this reason, the (*Degree Days API*) was used to import historical HDD and CDD values for this dataset by mapping the respective zip codes to their corresponding HDD and CDD values. Similar to the RECS dataset, this study applied 4 HDD and CDD variables – one for the year of analysis (2018) and one for the historical average. Using comprehensive exploratory data analysis (EDA), (Asamoah & Shittu, 2025) detected the pattern and correlation between various variables that influence heating and cooling loads. A methodology called Variance Inflation Factors (VIF) was adopted to prevent multicollinearity in the machine learning algorithm by dropping independent variables with high correlation. The next phase of their filtration process utilized a multi-task learning model that could concurrently forecast both cooling and heating requirements while identifying only the most critical features, effectively nullifying the coefficients of less impactful features. The final step was to use recursive feature elimination (RFE) to remove the least important features, thereby helping prevent overfitting and enhancing the model’s performance. As a result, (Asamoah & Shittu, 2025) were left with a final total of 19 features for model training. To be consistent with the analysis and to compare with the RECS dataset, the equivalent features in the ResStock dataset were selected for analysis, as shown in Table 2. Approximately 27 equivalent RECS characteristics remained. Furthermore, we identified 13 features from (Asamoah & Shittu, 2025) that didn’t overlap with the features selected for this analysis. This led to the inclusion of the 14 features from (Erickson et al., 2014)’s study to this study’s set of features and developed models for forecasting. The results of the analysis are discussed in Section 4.

*Table 2: List of features used for the ML model development for the ResStock Dataset*

| API_HDD_2018 | API_CDD_2018 | HDD_21y_average | CDD_21y_average |
|---|---|---|---|
| in.geometry_building_type_recs | in.vintage | in.bedrooms | in.windows |
| in.geometry_stories | in.infiltration | in.hvac_heating_type_and_fuel | in.heating_fuel |
| in.hvac_cooling_type | in.cooling_setpoint_has_offset | in.heating_setpoint_has_offset | in.water_heater_fuel |
| in.occupants | in.lighting | in.geometry_floor_area | in.ashrae_iecc_climate_zone_2004_2_a_split |
| in.insulation_wall | in.geometry_foundation_type | in.geometry_attic_type | in.cooking_range |
| in.dishwasher | in.clothes_dryer | in.refrigerator | in.clothes_washer |
| in.income | in.income_recs_2015 | in.income_recs_2020 | out.params.wall_area_above_grade_conditioned_ft_2 |
| out.params.window_area_ft_2 | out.params.roof_area_ft_2 | in.hvac_cooling_efficiency | in.hvac_heating_efficiency |
| in.hvac_cooling_partial_space_conditioning | in.heating_setpoint | in.cooling_setpoint | out.electricity.mech_vent.energy_consumption.kwh |
| out.electricity.summer.peak.kw | out.electricity.winter.peak.kw | | |

### 3.2.2. Data pre-processing

Similar to the RECS dataset, data pre-processing was carried out to ensure the data is clean and can be used for the smooth development of the machine learning models. A critical step in the ResStock dataset was the elimination of data points with missing HDD and CDD values. Another factor in the elimination of data points was the residential facility's occupancy status. For instance, if the facility was marked as vacant, the data point was eliminated from consideration for this study, similar to (Erickson et al., 2014). An additional criterion in the data preprocessing step for space heating and space cooling was to eliminate all samples that lacked primary heating and primary cooling equipment, respectively. For all three analyses, a logarithmic transformation approach was employed to eliminate negative predictions in the presence of "0 values", discussed further in Section 3.2.3. Once this step was completed, the subsequent stages, which included one-hot encoding of categorical variables, z-score normalization of numerical variables, and outlier removal through k-means clustering, followed a similar approach as outlined for the RECS dataset in Section 3.1.2. The remaining steps used for the ResStock dataset followed a similar pattern of analysis. Section 4 discusses the results obtained for each of the datasets and how they perform compared to studies in literature.

### 3.2.3. Logarithmic transformation of target variables

In modeling residential energy consumption and heating/cooling loads, the target variable often spans several orders of magnitude. Simple regression on the raw scale may result in negative predictions, even though the underlying quantities are inherently non-negative. To address both the lower-bound constraint (zero) and the right-skewed distribution typical of energy use, a logarithmic transformation was applied to the target variable. Specifically, the relationship $y^* = ln(y + c)$ was used, where $y^*$ represents the original consumption metric (in kWh or equivalent), and $c$ is a small positive constant added to shift zero-valued observations into the positive domain (since $ln(0)$ is undefined). This transformation offers three key advantages (F. Zhang et al., 2016): (1) it ensures that the exponentiated model output, $\hat{y} = exp(\hat{y}^*) - c$, remains strictly non-negative, thus preventing nonsensical negative forecasts; (2) it compresses the right tail of the distribution, which helps mitigate heteroscedasticity and reduces the impact of extreme high-consumption outliers on the loss function; and (3) it linearizes multiplicative relationships between predictors and consumption, enhancing the fit of linear (or tree-based) algorithms that operate under additive error assumptions. Caution is necessary when adding a constant, as it modifies the effective scale of the data. This adjustment requires careful back-transformation to mitigate biases from Jensen's inequality and retransformation errors (Kolter & Ferreira, 2011). Additionally, the choice of this constant can subtly impact predictions, especially at the lower end of the distribution. All models were trained using this log-transformed data, and the outputs were evaluated on the original scale by exponentiating and subtracting the constant. This methodology offers a systematic approach to prevent negative forecasts while preserving interpretability and enhancing model stability in the context of residential energy prediction.

# 4. Results and Discussion

## 4.1. RECS

### 4.1.1. Energy performance forecasting

This study evaluated several tree-based machine learning models for predicting energy performance and compared their performance, as shown in Table 3. CatBoost stands out as the best-performing model, achieving the highest $R^2$ value for all three target variables: total energy consumption, space heating, and cooling. Moreover, CatBoost's scores for Mean Absolute Error (MAE), Mean Squared Error (MSE), and Root Mean Squared Error (RMSE) are among the lowest, indicating lower prediction errors than those of the other models. This robust performance of CatBoost underscores its ability to effectively capture the complex relationships between various features and energy consumption.

*Table 3: Performance comparison of tested ML models for the RECS Dataset*

| Target variable | Model | $R^2$ | MAE (kWh) | RMSE (kWh) |
| --- | --- | --- | --- | --- |
| | LightGBM | 0.714 | 17,795 | 24,294 |
| Total Energy Consumption | Random Forest | 0.691 | 18,506 | 25,242 |
| | CatBoost | **0.726** | **17,338** | **23,788** |
| | XGBoost | 0.695 | 18,444 | 25,090 |
| | Neural Network | 0.718 | 17,750 | 24,134 |
| Space Heating | LightGBM | 0.728 | 11,115 | 16,472 |
| | Random Forest | 0.707 | 11,452 | 17,097 |
| | CatBoost | **0.732** | **11,001** | **16,348** |
| | XGBoost | 0.697 | 11,736 | 17,364 |
| | Neural Network | 0.728 | 11,127 | 16,459 |
| Space Cooling | LightGBM | 0.641 | 2,369 | 3,512 |
| | Random Forest | 0.631 | 2,387 | 3,565 |
| | CatBoost | **0.653** | **2,346** | **3,455** |
| | XGBoost | 0.609 | 2,487 | 3,669 |
| | Neural Network | 0.628 | 2,481 | 3,576 |

Table 3 illustrates the performance of the evaluated models across three target variables, utilizing various error metrics. In terms of total energy consumption, CatBoost demonstrates the strongest overall performance, achieving the highest coefficient of determination ($R^2 = 0.726$) and the lowest MAE (17,337 kWh) and RMSE (23,788 kWh). This indicates a consistent improvement in predictive accuracy relative to the other models. A similar trend is observed for space heating and space cooling, where CatBoost again offers the best or near-best performance across most metrics. However, the differences in performance among the models are relatively minor. These results imply that the implemented feature engineering approach and hyperparameter tuning yield incremental yet consistent enhancements across varying prediction tasks. Compared to the findings of Korsavi et al. (2025), the proposed models showcase a modest improvement in predictive performance for total energy consumption ($R^2$ = 0.726 vs. 0.712). Although this enhancement is incremental, it underscores the significance of meticulous model development practices in boosting machine learning performance. It is also noteworthy that due to near-zero target values, MAE and RMSE are used as more reliable metrics for comparison in these instances.

**Regression Analysis**

In regression analysis, actual-versus-predicted plots are used to evaluate model performance by comparing predicted outputs with observed values from the dataset (*Hands-On Machine Learning with Scikit-Learn, Keras, and TensorFlow, 2nd Edition*). A well-performing model is characterized by points clustering closely around the 1:1 diagonal line, indicating strong agreement between predictions and observations. Systematic deviations from this line may reveal model deficiencies, such as bias, heteroskedasticity, or an inability to capture non-linear relationships (Cao et al., 2020). Beyond aggregate performance metrics, this visual diagnostic provides additional insight into model behavior by enabling direct assessment of prediction errors across the response range. In particular, it facilitates the identification of unexpected patterns, outliers, and variance for specific subsets of the data that may not be evident from numerical metrics alone. In this study, such plots are used to complement quantitative evaluation metrics and to assess the robustness of the developed models across different energy performance targets.

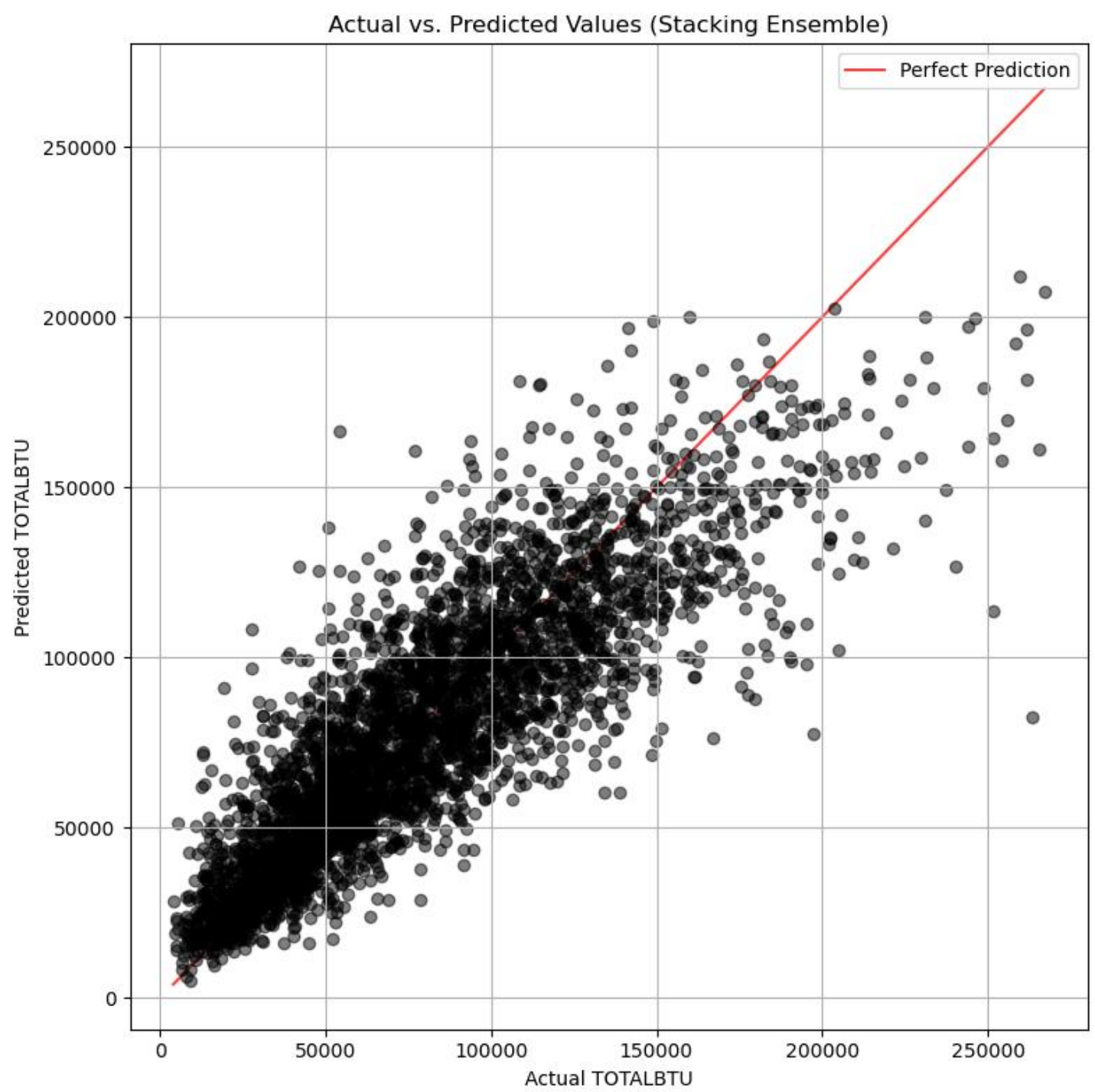

*Figure 2: Regression of the actual vs predicted total energy consumption values for the CatBoost Machine Learning model for the RECS dataset*

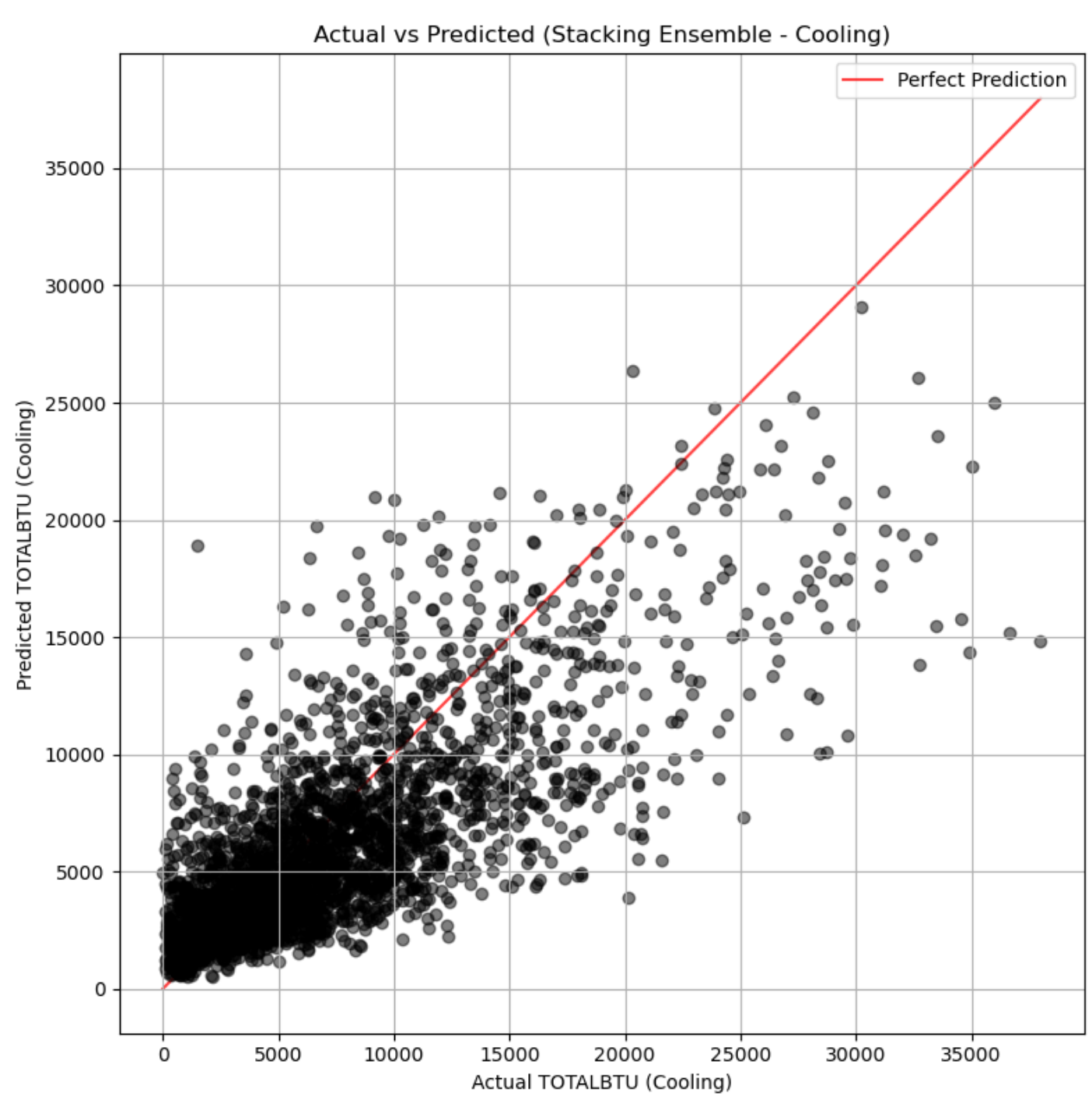

*Figure 4: Regression of the actual vs predicted space cooling values for the CatBoost Machine Learning model for the RECS dataset*

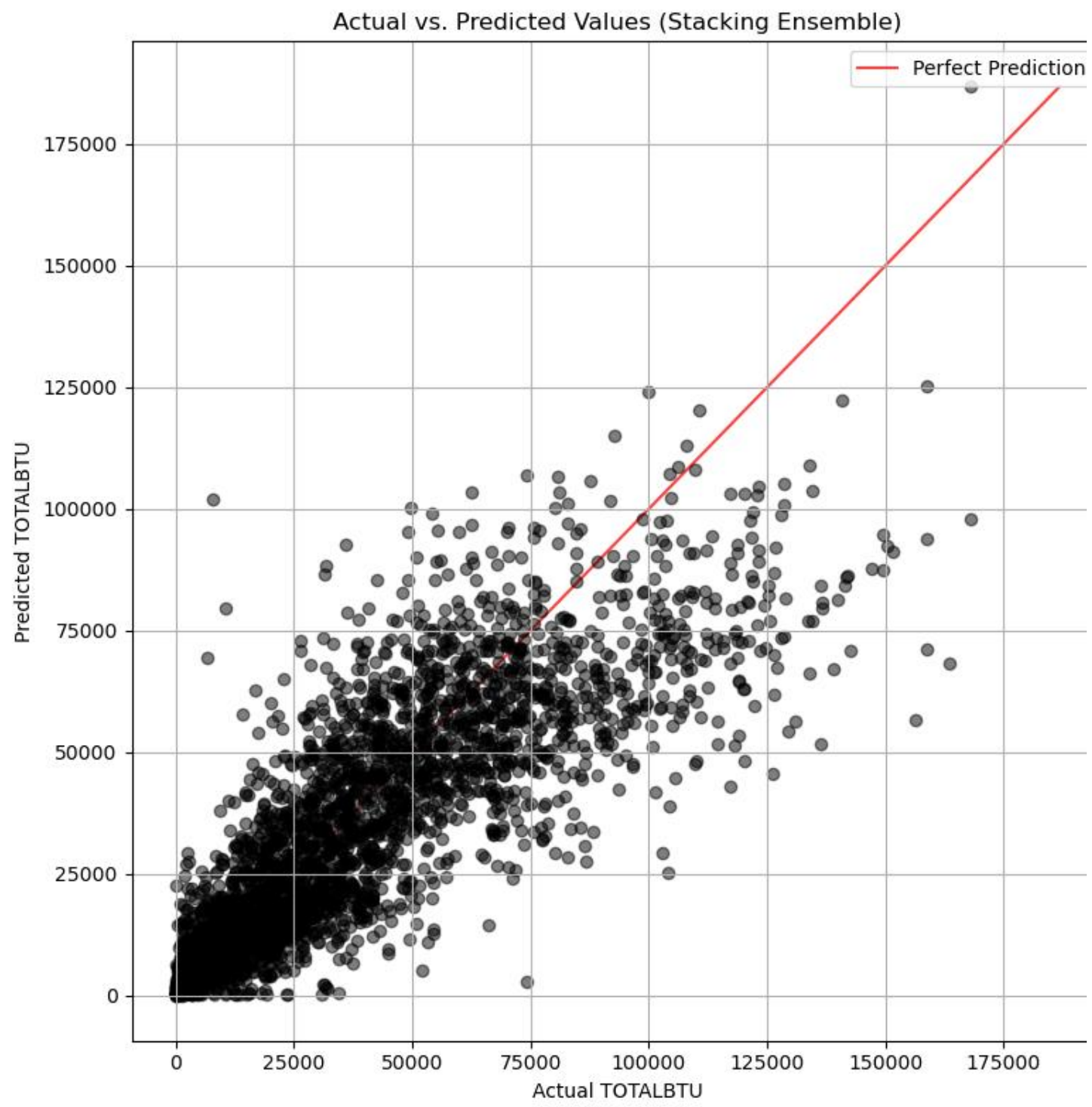

*Figure 3: Regression of the actual vs predicted space heating values for the CatBoost Machine Learning model for the RECS dataset*

The three scatter plots presented here, Figures 2-4, compare actual and predicted values for total energy consumption (Figure 2), space heating (Figure 3), and space cooling (Figure 4) using a stacking ensemble model built upon a finely tuned CatBoost model. Each plot includes a diagonal red line representing the "perfect prediction" scenario, where the predicted values align precisely with the actual values. In Figure 2, which illustrates total energy consumption, the data points exhibit a strong linear relationship, with the predictions closely matching the actual values. However, some scattering is evident as values increase, indicating minor underestimations at higher consumption levels. Figure 3, focused on space heating, shows a similar trend, though the scatter is more pronounced, reflecting greater variability between predicted and actual values. As with total energy consumption, the model aligns reasonably well with actual values at lower levels of space heating, but discrepancies become more significant at higher levels, suggesting that the model's performance is somewhat compromised by data extremes. Lastly, Figure 4, which addresses space cooling, reveals an even greater divergence, especially at higher values, where the predicted values consistently overestimate actual cooling needs. This discrepancy may underscore the challenges of accurately modeling cooling requirements, particularly in buildings with high cooling loads. Overall, these plots suggest that while the stacking ensemble model performs well in general, there are areas for improvement in capturing extremes in energy consumption, particularly for space heating and cooling. The model's predictive power appears strongest for lower levels of energy consumption across all categories.

When analyzing energy performance with stacking ensemble models, comparing predicted energy consumption to actual values in a scatter plot reveals important insights into model accuracy. Significant vertical deviations from the ideal 1:1 line indicate consistent over- or under-prediction, suggesting biases related to specific building types or conditions. The spread of data points indicates predictive uncertainty: a tighter cluster indicates better performance, while wider dispersion suggests higher variance due to unaccounted factors like occupant behavior or equipment inefficiencies (Vinson Joshua et al., 2022). Outliers, points far from the regression line, may indicate serious prediction errors or unusual data that require careful investigation to determine their cause, whether due to data issues or unique operational circumstances. Understanding these deviations and outliers is crucial for improving model calibration, quantifying uncertainty, and strengthening the reliability of building energy management decisions.

### 4.1.2. Drivers of energy performance – SHAP sensitivity analysis

Utilizing the most reliable model for energy performance forecasts from Table 3 (CatBoost), a SHAP sensitivity analysis was conducted to identify the key drivers influencing energy consumption. Figures 5-7 present waterfall and beeswarm plots that illustrate the most influential features, showcasing the distribution of their SHAP values across individual samples. The features are ranked according to their aggregate impact on the model output. The horizontal axis shows the SHAP value, and the color gradient, from blue to red, indicates the feature value, with red indicating higher values. Features located on the right side of the graph, characterized by higher SHAP values and shown in red, are associated with increased energy consumption. This indicates that a rise in their values or presence is associated with higher energy use. Similarly, features on the left side of the graph with higher SHAP values, shown in blue, reduce energy consumption, meaning their increased values or presence are linked to lower energy usage.

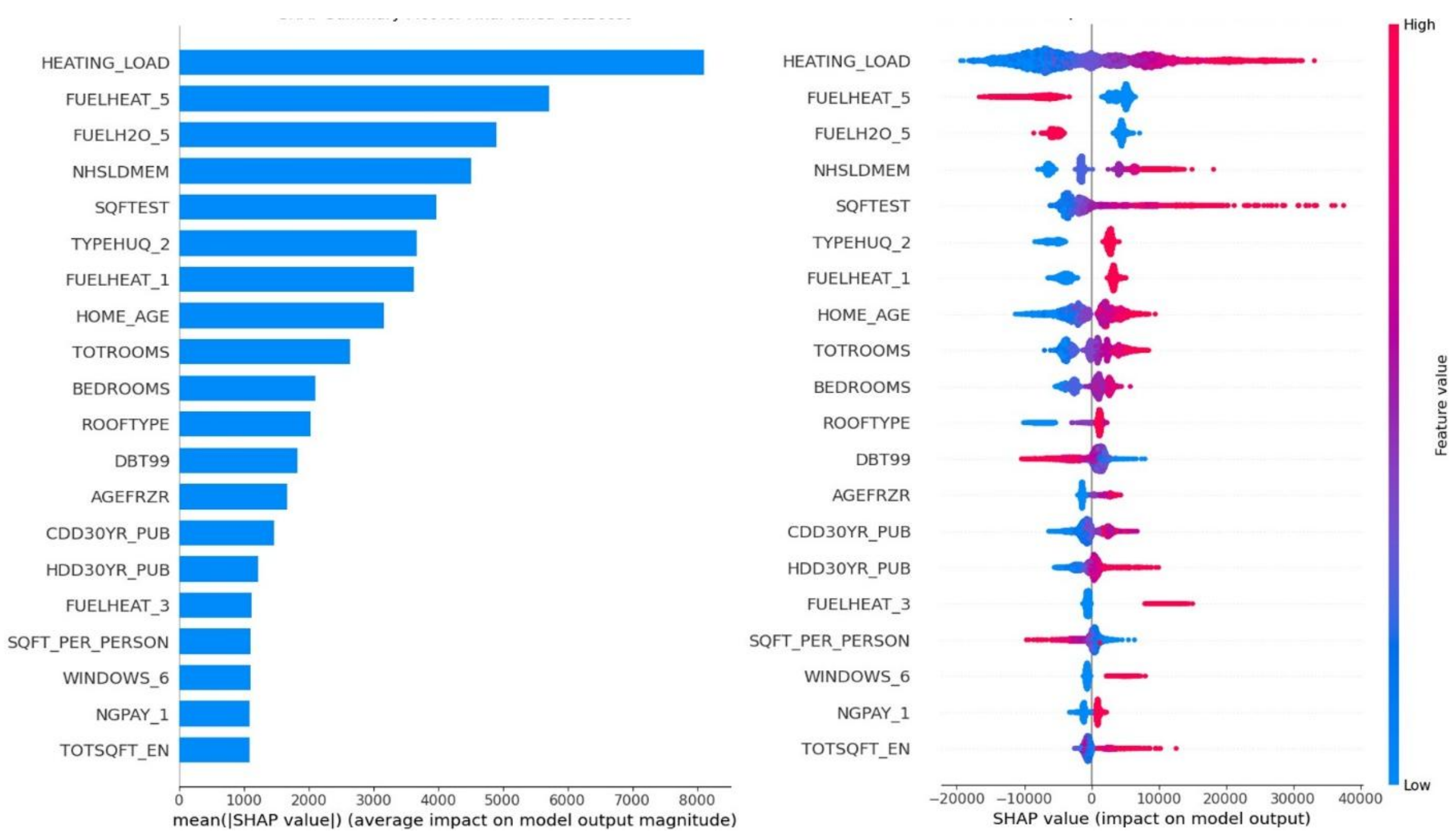

*Figure 5: SHAP values for the most correlated features across all samples for total energy consumption for the RECS Dataset*

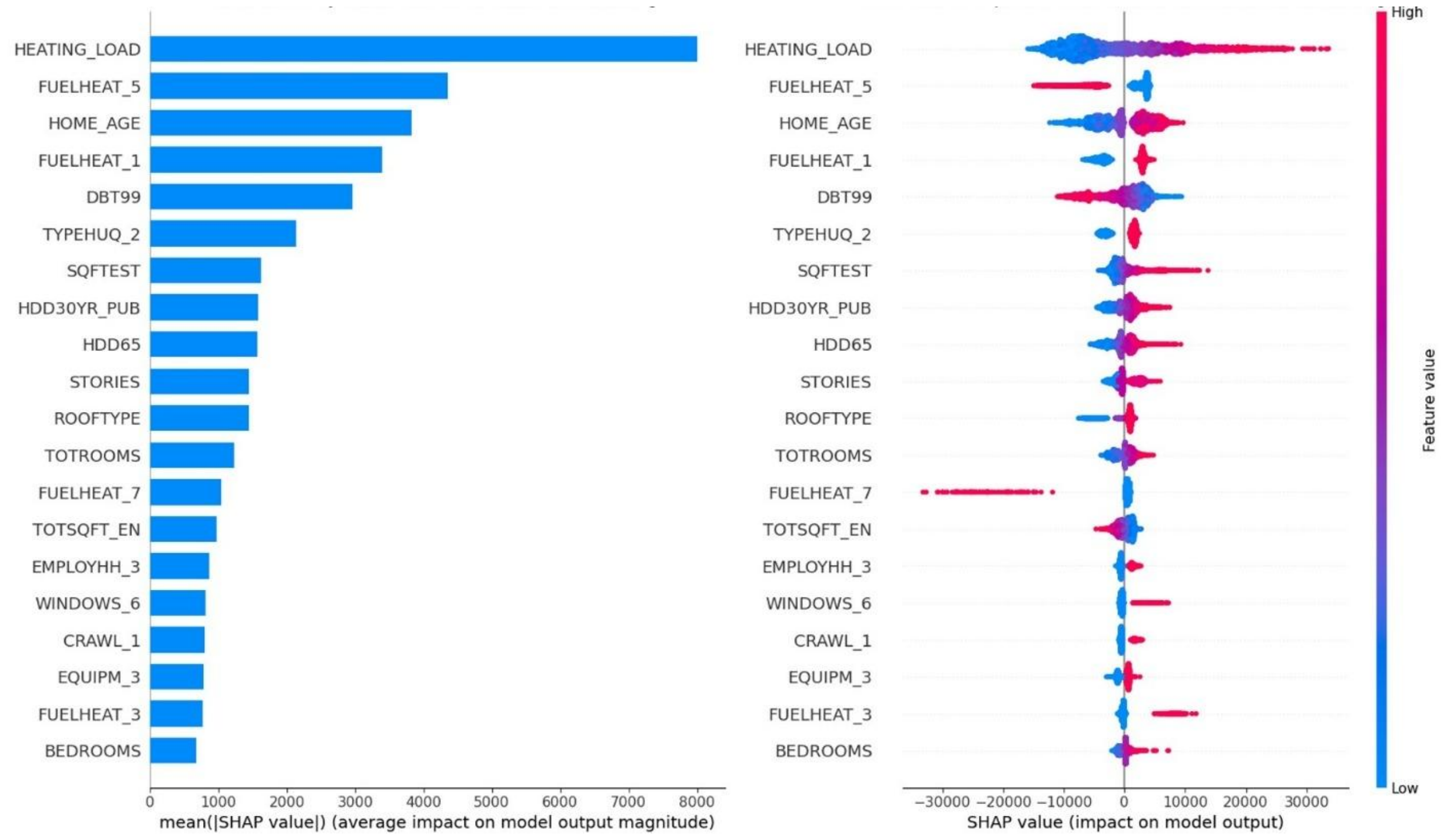

*Figure 6: SHAP values for the most correlated features across all samples for space heating for the RECS Dataset*

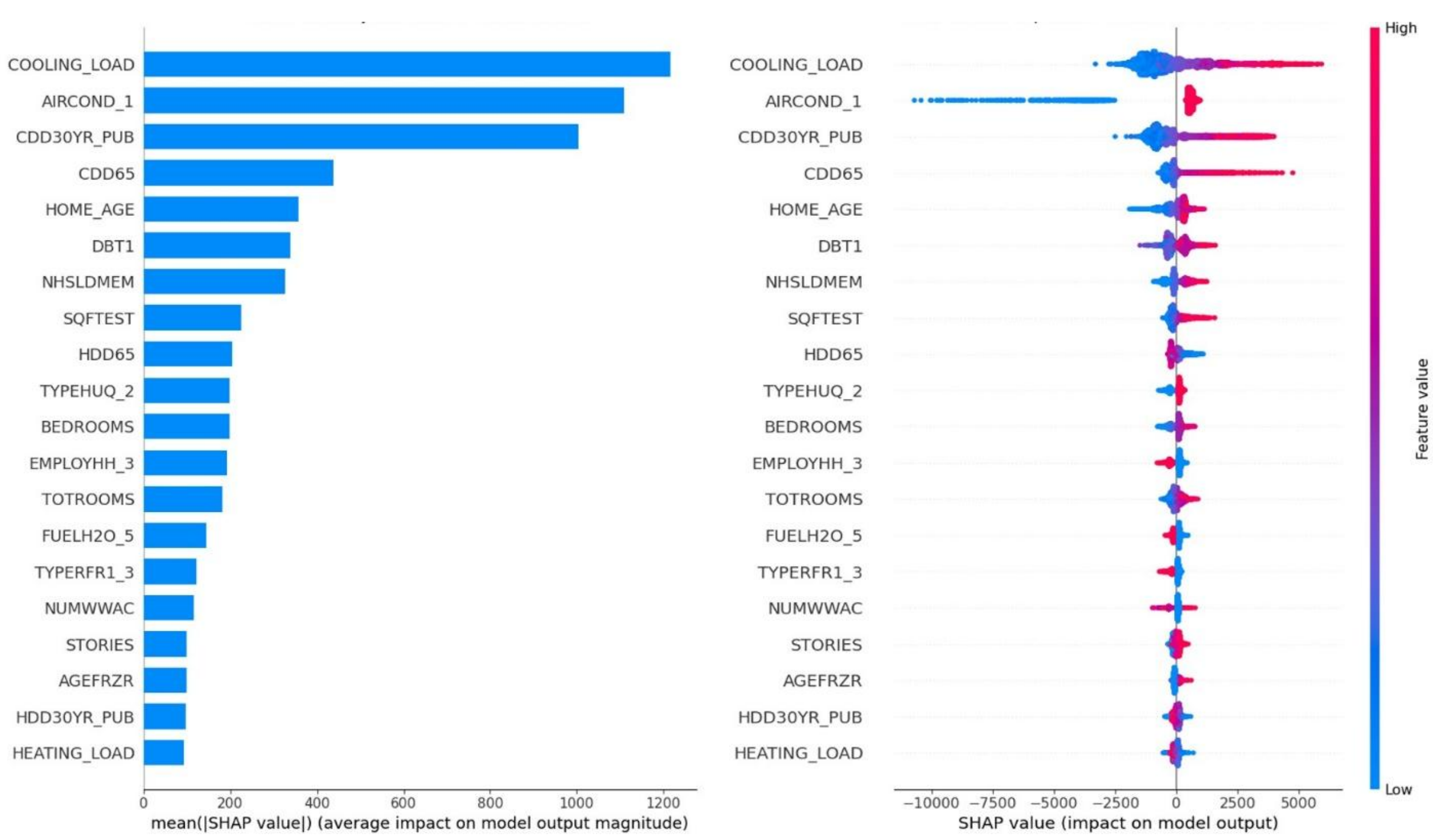

*Figure 7: SHAP values for the most correlated features across all samples for space cooling for the RECS Dataset*

In the first analysis for total energy consumption prediction, using a tuned CatBoost model, HEATING_LOAD was identified as the most influential predictor for total energy consumption, followed by fuel type indicators (FUELHEAT_5 (electricity), FUELH2O_5 (electricity), FUELHEAT_1 (natural gas), household characteristics (NHSLDMEM for number of household members, SQFTEST for square footage, and HOME_AGE), along with structural attributes like TOTROOMS and ROOFTYPE. High heating loads and specific fuel types consistently led to increased total energy use. The second analysis using the CatBoost model assessed space-heating energy. Here, HEATING_LOAD again dominated, with critical drivers like HOME_AGE, FUELHEAT_5 (electricity), FUELHEAT_1 (natural gas), and weather variables (DBT99 (Dry Bulb Design Temperature – expected to exceed 99% of the time) and HDD30YR_PUB (Heating degree days: 30-year average)). These findings confirm that older homes and colder regions are associated with higher heating energy use. In the third analysis, the tuned CatBoost model examined space cooling energy use, identifying COOLING_LOAD as the key factor, followed by AIRCOND_1 (air conditioning presence) and weather-related variables (CDD30YR_PUB (Cooling degree days: 30-year average) and CDD65). Secondary drivers included HOME_AGE, DBT1 (Dry Bulb Design Temperature – expected to be exceeded 1% of the time), and SQFTEST. The analysis showed that homes with higher cooling loads and existing air conditioning systems in warmer regions consume more cooling energy, consistent with trends observed in measured data and established building energy studies, thereby demonstrating that the model reproduces expected real-world behavior. Overall, SHAP analysis provided a clear understanding of the factors influencing residential energy consumption by ranking feature importance and illustrating their impact on predictions. A key thing to note in Figures 5-7 is the prevalence of the 3 out of the 5 engineered features, i.e., 'HEATING_LOAD', 'HOME_AGE', and 'SQFT_PER_PERSON', as some of the key determinants influencing energy consumption.

## 4.2. ResStock

### 4.2.1. Energy performance forecasting

This section delves into the training and evaluation of seven distinct machine learning models aimed at predicting the total energy consumption as well as space heating and cooling of residential buildings. The same five machine learning algorithms used in the RECS analysis were also employed to determine which models effectively predict these loads. The algorithms analyzed include Light Gradient Boosting (LightGBM), Random Forest, CatBoost, XGBoost, and Neural Networks. Each model was evaluated based on its learning performance and its capacity to minimize errors during training. The same three accuracy metrics: RMSE, MAE, and $R^2$ were used to compare the individual predictive performances of the models, results of which are shown in Table 4.

*Table 4: Performance comparison of tested ML models for the ResStock dataset*

| Target variable | Model | $R^2$ | MAE (kWh) | RMSE (kWh) |
|---|---|---|---|---|
| Total Energy Consumption | LightGBM | 0.795 | 5,397 | 7,462 |
| | Random Forest | 0.865 | 4,346 | 6,053 |
| | CatBoost | **0.905** | **3,636** | **5,077** |
| | XGBoost | 0.884 | 4,096 | 5,621 |
| | Neural Network | 0.899 | 3,815 | 5,253 |
| Space Heating | LightGBM | 0.832 | 11,477 | 16,838 |
| | Random Forest | 0.895 | 8,566 | 13,326 |
| | CatBoost | **0.927** | **7,038** | **11,145** |
| | XGBoost | 0.916 | 7,961 | 11,906 |
| | Neural Network | 0.922 | 7,264 | 11,485 |
| Space Cooling | LightGBM | 0.845 | 7,088 | 10,161 |
| | Random Forest | 0.899 | 5,455 | 8,213 |
| | CatBoost | **0.930** | **4,455** | **6,809** |
| | XGBoost | 0.911 | 5,236 | 7,676 |
| | Neural Network | 0.921 | 4,843 | 7,254 |

Similar to the RECS analysis, the CatBoost model achieved the best performance across all three target applications. For the ResStock dataset, CatBoost attained $R^2$ values of 0.905, 0.927, and 0.930 for total energy consumption, space heating, and space cooling, respectively, consistently outperforming the other models across all error metrics. These values represent a clear improvement over those obtained from the RECS dataset, indicating enhanced predictive accuracy for the ResStock-based models. This performance gain can be attributed to several factors. Unlike RECS, which is survey-based and subject to reporting uncertainty and coarser variable resolution, the ResStock dataset is generated through physics-based simulations, providing more consistent and higher-resolution input features. This likely enables machine learning models to better capture underlying relationships between building characteristics and energy consumption. CatBoost excels in machine learning regression, particularly with categorical data, due to its built-in handling of categorical features and its innovative training approach that prevents target leakage and overfitting (Hancock & Khoshgoftaar, 2020). Unlike many other algorithms that require manual preprocessing, CatBoost automatically and effectively processes categorical variables, saving significant time and effort.

### 4.2.2. Regression Analysis

To gauge the spread of variances in the predicted values for total energy consumption as well as space heating and cooling, actual vs predicted scatter plots are shown in Figures 8-10. For total energy consumption, the data show good predictions, with uniform scatter along the red line indicating perfect prediction (y = x), as shown in Figure 8.

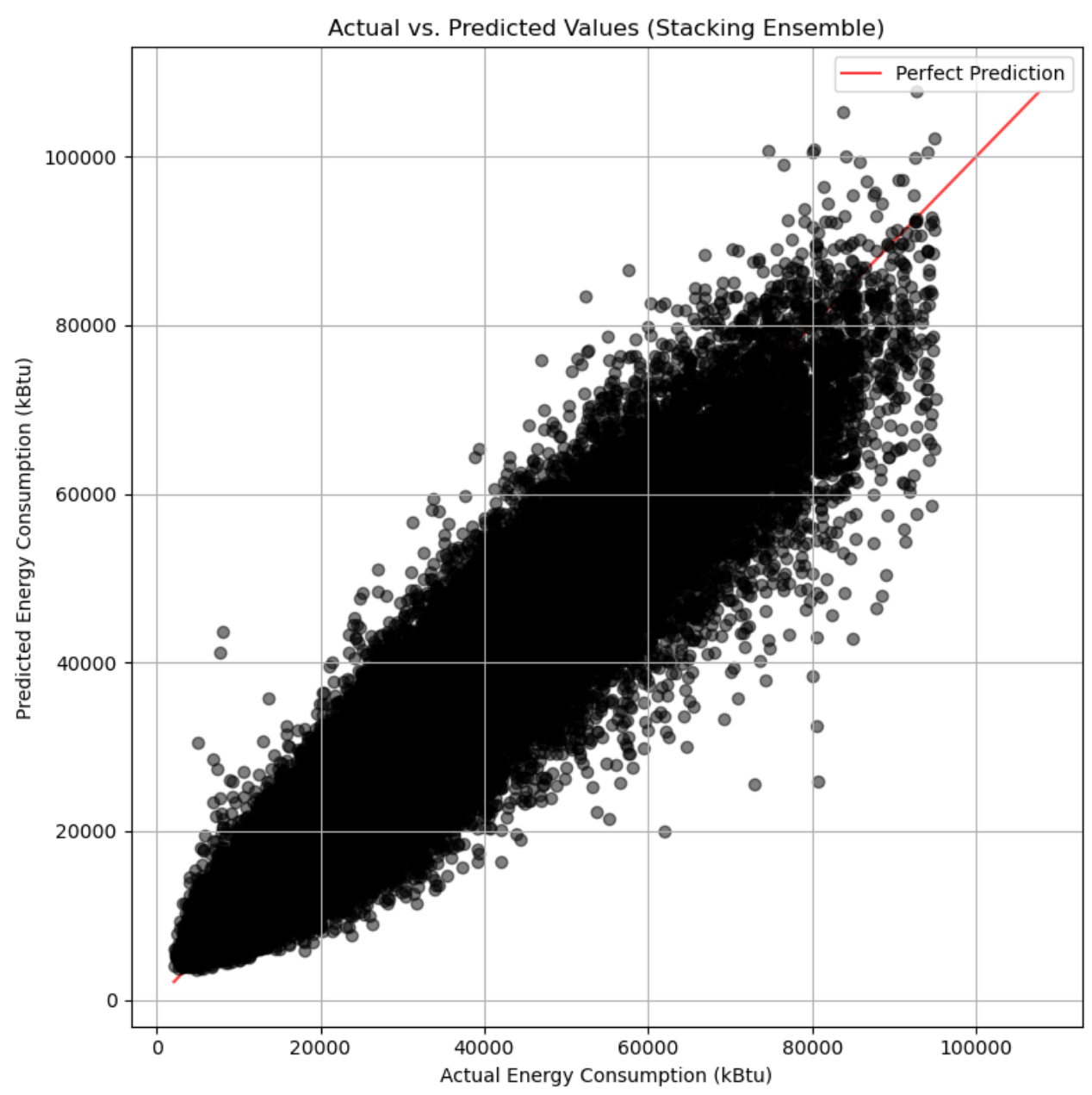

*Figure 8: Regression of the actual vs predicted total energy consumption values for the CatBoost Machine Learning model for the ResStock dataset*

Again, CatBoost was the best-performing model for the regression analysis for all three target variables, i.e., total energy consumption as well as space heating and cooling. A stacking ensemble technique, which enhances predictive accuracy through a hierarchical architecture is applied. This approach utilizes the diverse learning biases of multiple base algorithms to reduce localized variance and systemic error. The ensemble features a two-tier architecture: Level-0 includes independent models like Gradient Boosted Trees, Random Forests, and Support Vector Regressors, each capturing distinct patterns in energy consumption. At Level-1, a meta-learner processes cross-validated predictions from Level-0 to assign optimal weights, effectively balancing each model's strengths and weaknesses. Performance validation shows that this stacking technique improves regression accuracy, as evidenced by the tight clustering of data points along the identity line (y = x) and a high coefficient of determination ($R^2$). By integrating diverse hypothesis spaces, the ensemble mitigated overfitting and adeptly captured the complex nature of energy demand, confirming its superior generalization and high-precision energy forecasting capabilities. Figures 9 and 10 show the regression analysis for space heating and space cooling, respectively.

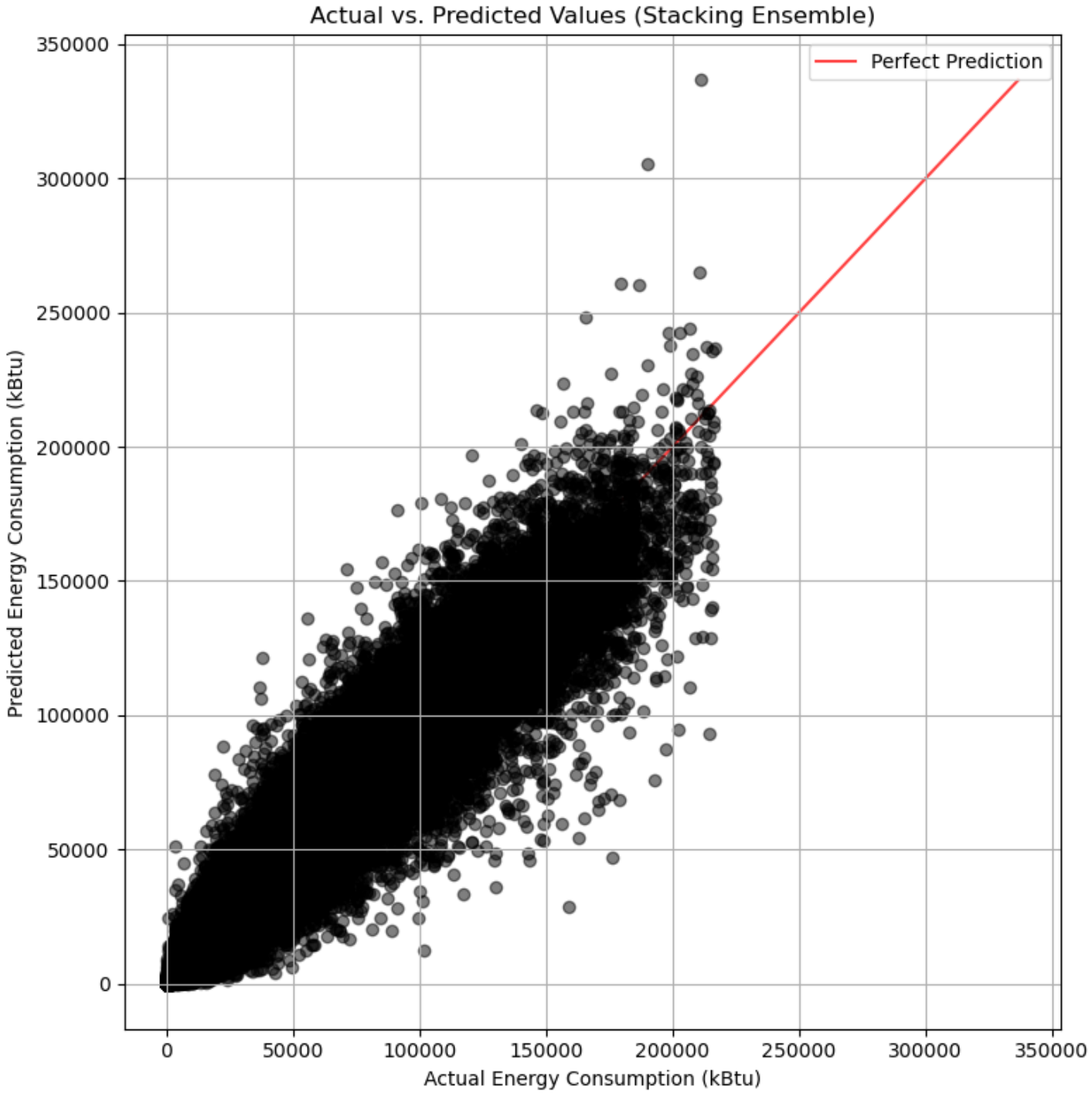

*Figure 9: Regression of the actual vs predicted space heating values for the CatBoost Machine Learning model for the ResStock dataset*

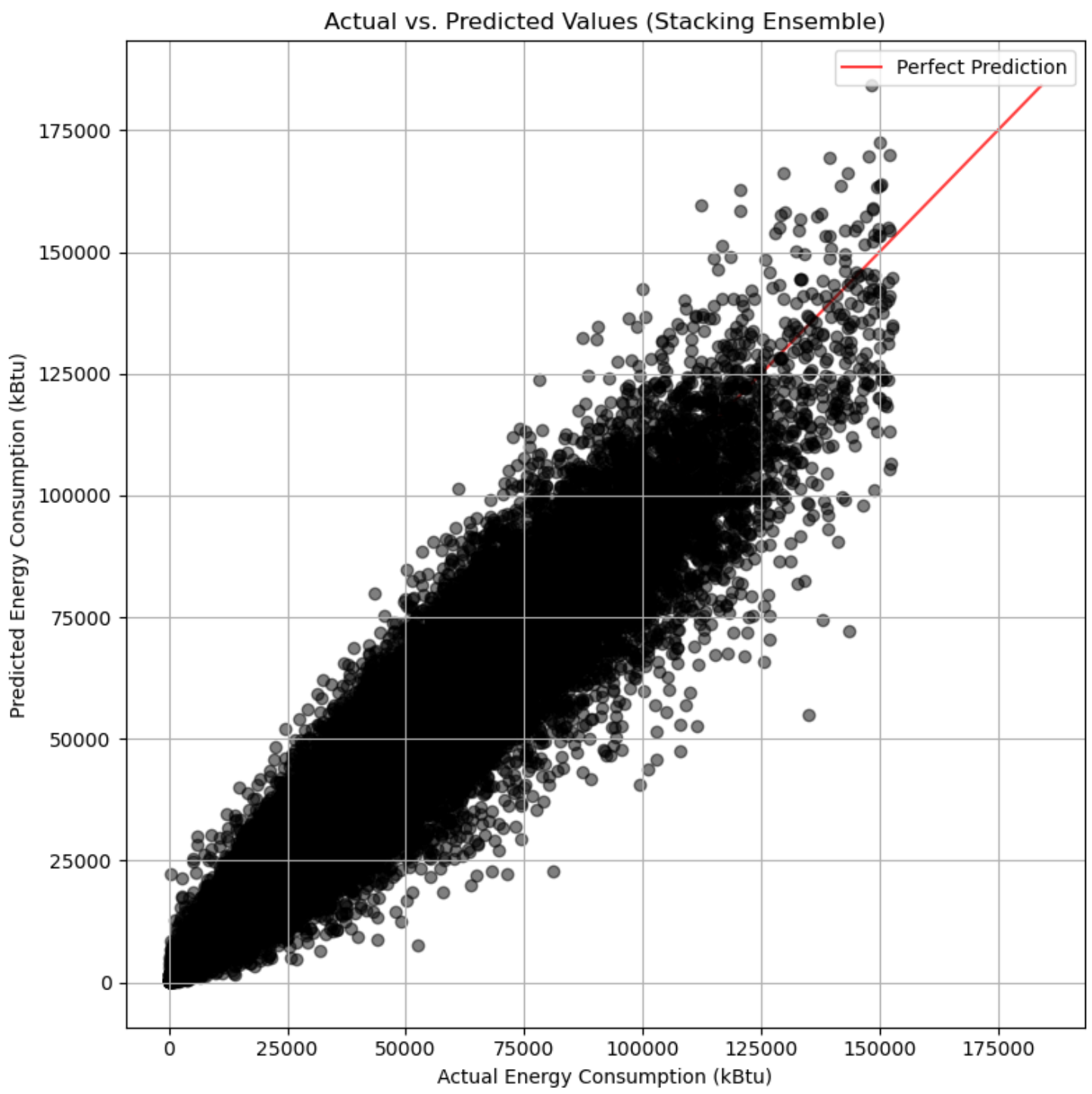

*Figure 10: Regression of the actual vs predicted space cooling values for the CatBoost Machine Learning model for the ResStock dataset*

In Figures 9 and 10, multiple datapoints were identified where the predicted values were either zero or negative. A review of the ResStock database revealed that there were indeed several instances in which the energy consumption for both space cooling and space heating was recorded as zero (55,745 for space cooling and 12,189 for space heating). Consequently, it was deemed appropriate to exclude these data points from consideration, particularly those where both consumptions were zero simultaneously, as they effectively represent a vacant facility.

#### 4.2.3. Drivers of energy performance

The SHAP sensitivity analyses for total energy consumption, space heating, and space cooling (Figures 11-13) reveal distinct yet complementary patterns regarding the drivers of residential energy performance. Across all energy end uses, weather factors and envelope geometry emerge as key determinants. In the case of total energy consumption, heating degree days (API_HDD_2018) and conditioned wall area exhibit the highest SHAP magnitudes, indicating a strong relationship between thermal exposure, building size, and overall energy demand, with additional variability introduced by system-level variables such as natural gas heating and ventilation loads. Space heating demand is predominantly influenced by HDD, envelope area, and insulation quality, while high infiltration rates and older building vintages exacerbate heating loads, emphasizing the sensitivity of heating demand to both conductive and convective losses. Conversely, space cooling is mainly driven by summer peak electricity loads and cooling degree days (API_CDD_2018), followed by solar-exposed envelope surfaces and internal setpoint controls, with infiltration playing a lesser role. HVAC system characteristics and partial space conditioning levels further influence cooling demand. These findings collectively highlight that energy use in residential buildings is primarily governed by weather conditions (especially temperature) and envelope characteristics, while operational and system-level factors play important but secondary roles. This suggests that prioritizing envelope retrofits and passive design measures could effectively reduce energy demand across various end uses.

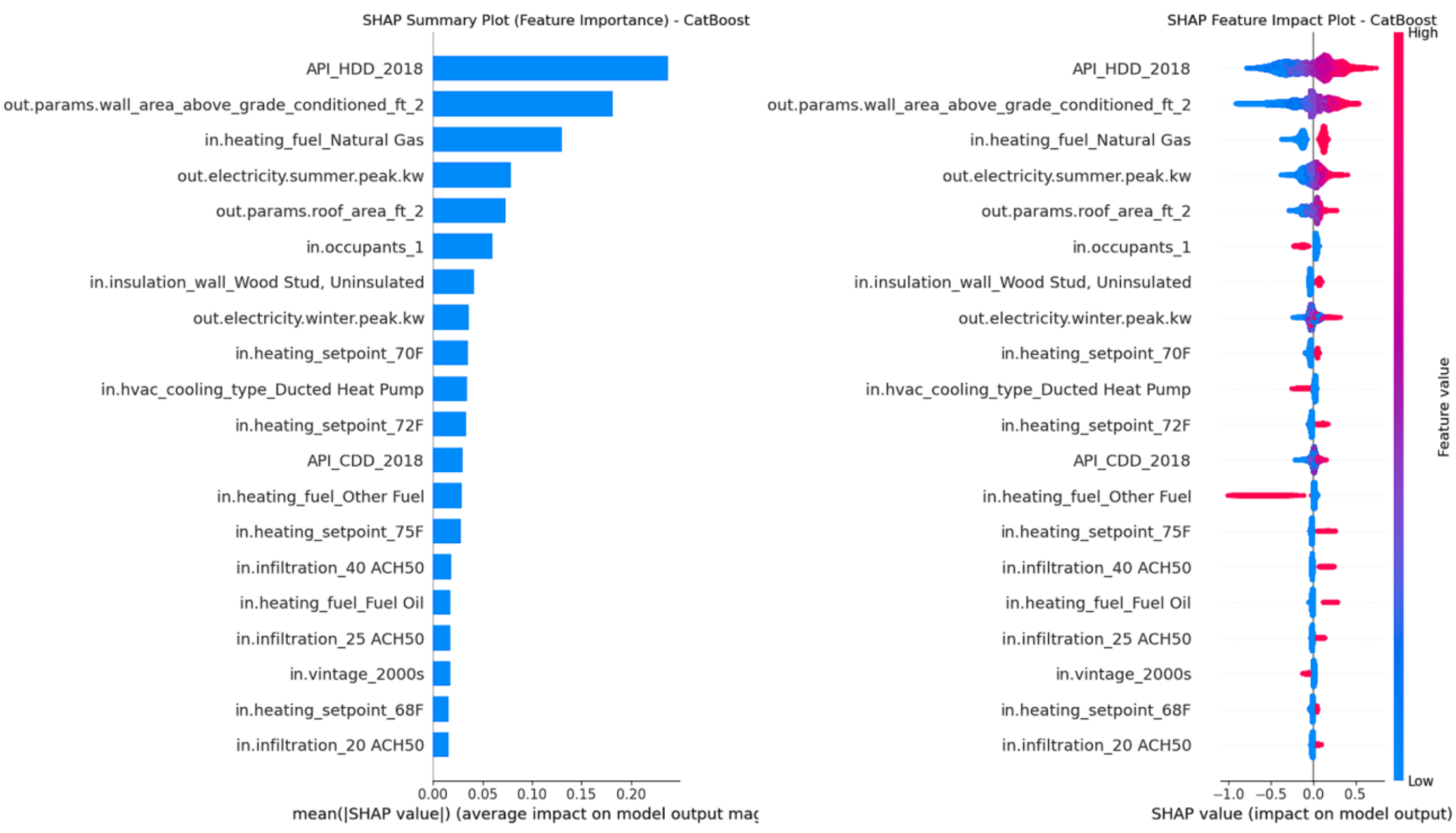

*Figure 11: SHAP values for the most correlated features across all samples for total energy consumption for the ResStock Dataset*

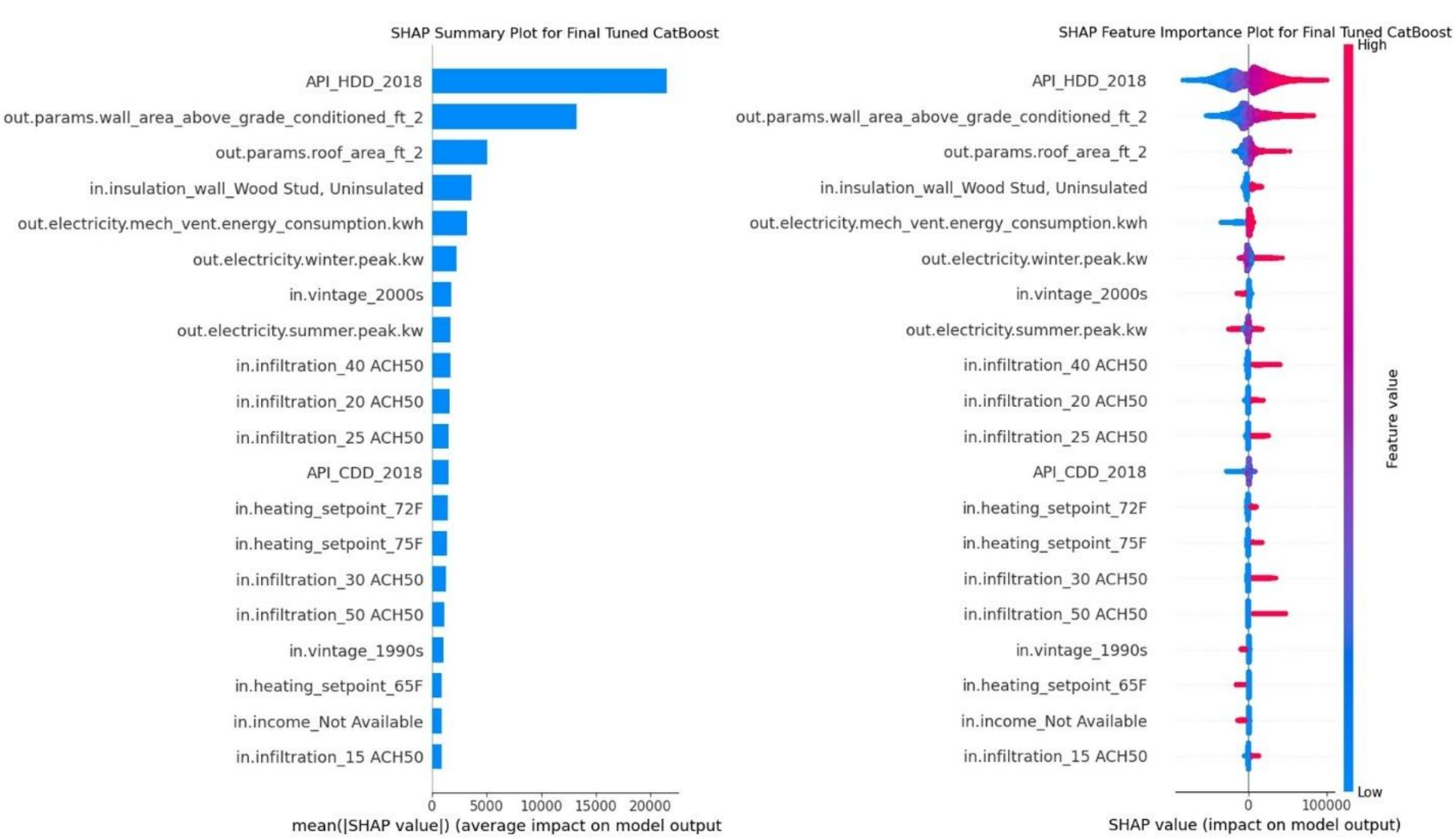

*Figure 12: SHAP values for the most correlated features across all samples for space heating for the ResStock Dataset*

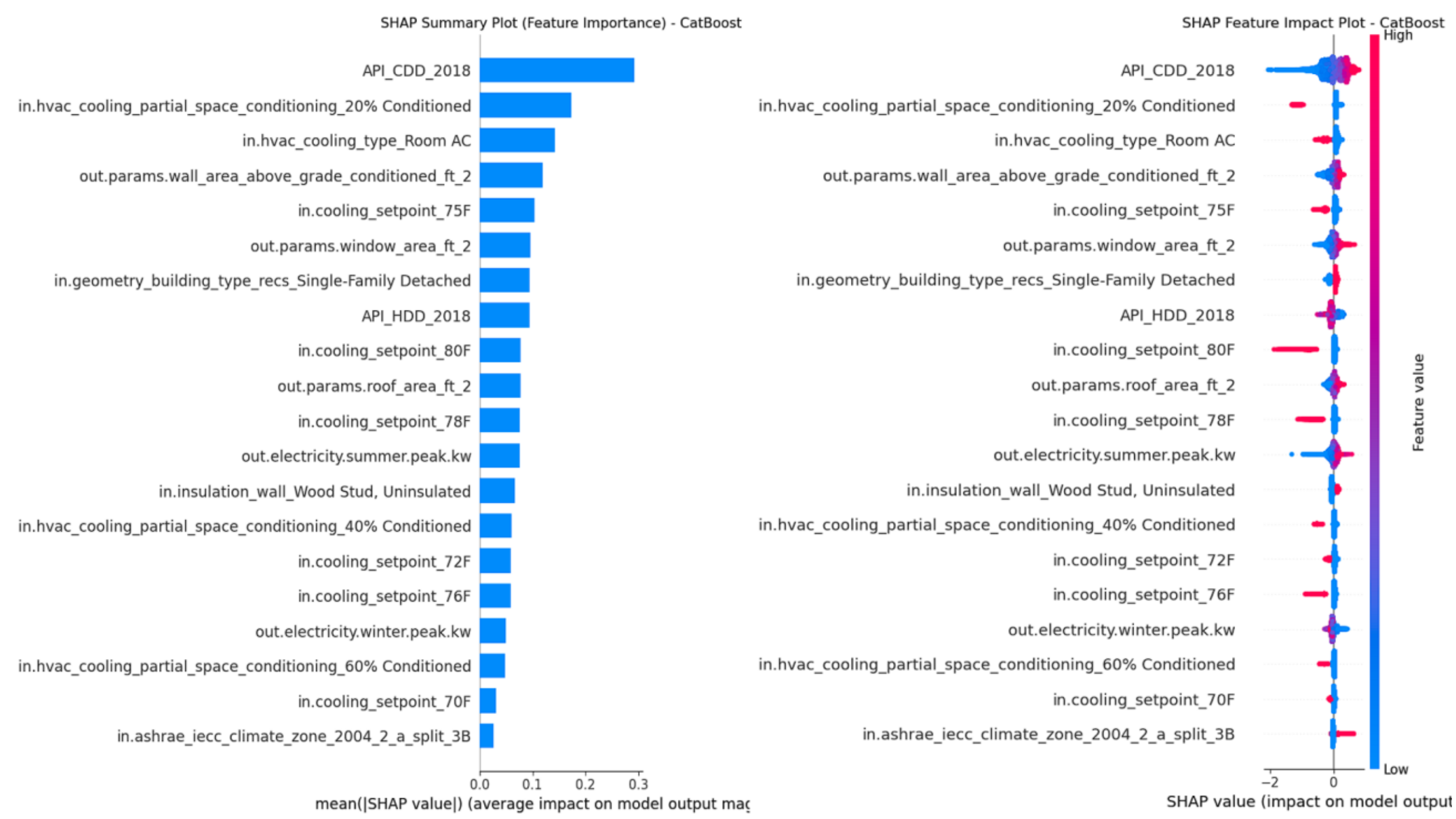

*Figure 13: SHAP values for the most correlated features across all samples for cooling for the ResStock Dataset*

### 4.3. Comparing RECS vs ResStock

The primary distinction between the RECS and ResStock datasets lies in their data collection methodologies. RECS is grounded in real-world surveys and billing information from actual households. Its data collection occurs in two phases: first, a household survey captures details on structural characteristics and energy-related behaviors; second, billing data is gathered from the nearly 18,500 households' energy providers. This empirical data-gathering initiative has resulted in the creation of RECS. In contrast, ResStock is a physics-based, statistically representative residential building-stock model developed by NLR. It combines housing-stock data with sampled dwelling-unit models and simulations in OpenStudio and EnergyPlus to generate detailed energy-use and retrofit-impact results across the U.S. housing stock.

These two datasets can be viewed as complementary rather than in competition (Glasgo et al., 2020). The RECS dataset provides observational, ground-truth data to validate the assumptions and accuracy of the ResStock model. Additionally, RECS offers valuable insights into historical and current energy trends based on actual consumption, while ResStock can forecast the future effects of emerging technologies and energy policies. While RECS provides national and, more recently, state-level statistics, ResStock can generate data at more granular levels, such as by census division, county, or even specific metropolitan areas. The primary objective of this study is to utilize two datasets to enhance our understanding and modeling of energy consumption within the U.S. residential sector.

The selection of input variables for the machine learning models was informed by standardized methodologies used to estimate residential utility allowances, which rely on a defined set of building and environmental characteristics to approximate energy performance. Consequently, ten user inputs were identified based on their relevance to the model, general accessibility for end-users, and their presence in both the RECS and ResStock datasets. While both RECS and ResStock provide valuable data for residential energy modeling, direct access to raw input data is often restricted due to privacy protections, proprietary data sources, or synthetic data generation methods (Geissler et al., 2019). Thus, only variables that are either publicly available or indirectly inferable were considered. Table 5 summarizes the final list of input features used for model training across both datasets.

*Table 5: List of inputs fed into the reduced-input model*

| Inputs |
|---|
| Heating Degree Days |
| Cooling Degree Days |
| Heating Degree Days (Historical average) |
| Cooling Degree Days (Historical average) |

| Type of housing (single family, multi-family, etc.) |
|---|
| Number of bedrooms |
| Fuel used by range |
| Main space heating fuel type |
| Air conditioning equipment used |
| Fuel used by main water heater |

Table 6 shows performance metrics obtained by applying the 10 inputs listed in Table 3 to the high-performing CatBoost model developed previously, and compares them with the same metrics obtained using RECS and ResStock data, respectively. As expected, the accuracy metrics for the 10 inputs for both datasets, were lower than before, as seen in Table 6, likely due to the reduced number of features used in prediction.

*Table 6: Comparison of the performance of the CatBoost Machine Learning Model for the ten inputs for the RECS and ResStock datasets*

| Dataset | $R^2$ | MAE (kWh) | RMSE (kWh) |
|---|---|---|---|
| **RECS** | 0.61 | 19,996 | 26,904 |
| **ResStock** | 0.62 | 7,226 | 10,057 |

Figures 14 and 15 together illustrate the interpretability-performance trade-off associated with using a reduced feature set that excludes some of the most impactful building features for predicting residential energy consumption in the ResStock dataset. Figure 14 displays a SHAP summary plot for the final tuned CatBoost model, highlighting the relative importance of the top 10 selected features.

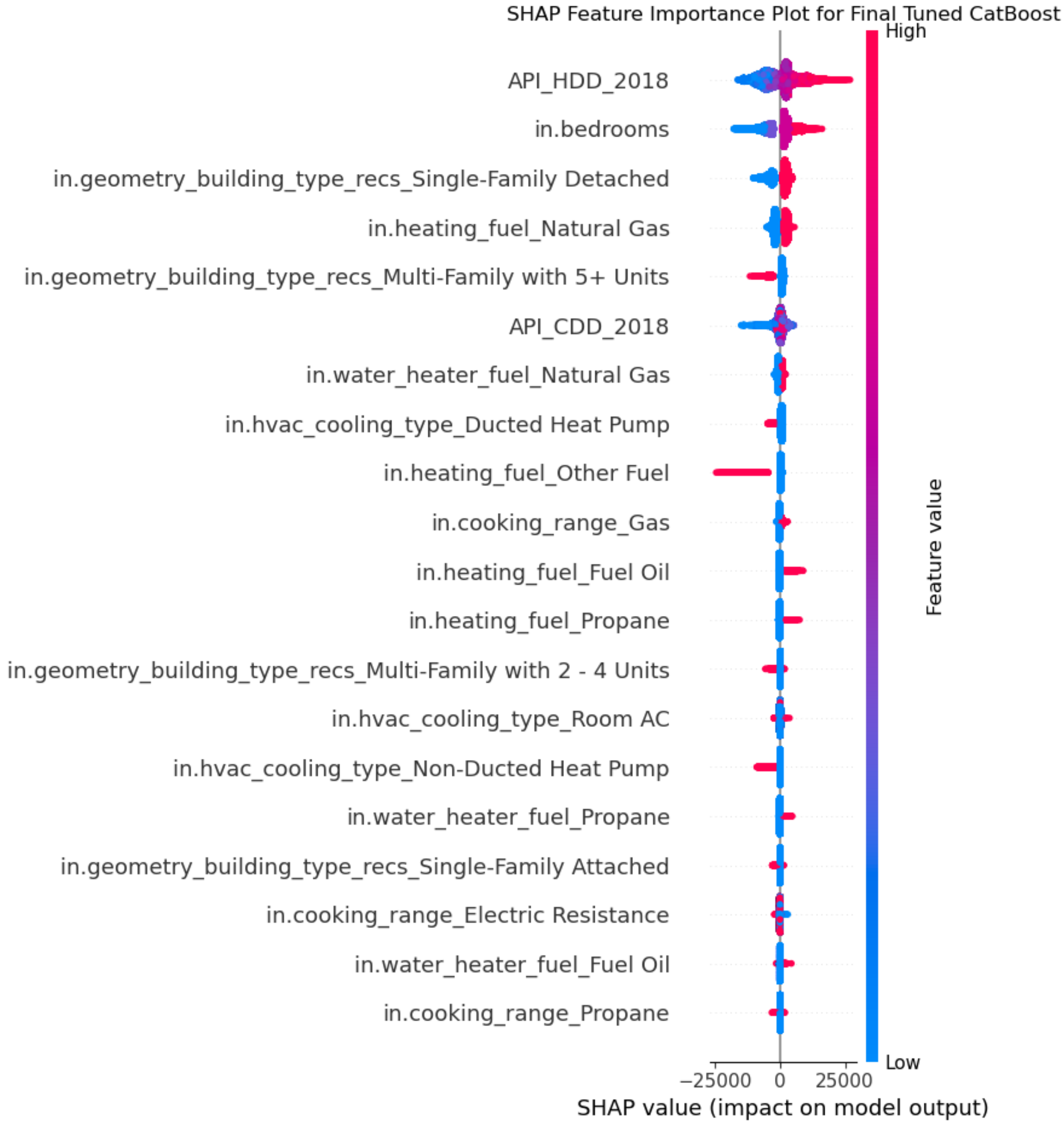

*Figure 14: SHAP sensitivity analysis for the ten inputs from Table 5 using the ResStock dataset and the developed CatBoost Machine Learning Model*

Key climatic indicators, such as 'API_HDD_2018' (heating degree days) and `API_CDD_2018` (cooling degree days), along with housing characteristics like the number of bedrooms and building type, as well as primary heating fuel, emerge as significant drivers of energy demand, as observed by their long-tailed distributions in the SHAP plot. Additionally, variations in heating and water heating fuel types also exert considerable influence. Absent from the pick list are building features in the RECS and ResStock datasets that demonstrate high correlation with energy use, such as wall area of conditioned above-grade space, roof area, and insulation levels. Figure 15 depicts an actual vs. predicted scatter plot for the stacking ensemble CatBoost model of the new runs using Table 5 inputs, revealing a noticeable deviation from the perfect prediction line, particularly at higher levels of energy consumption.

This observation is quantitatively supported by a moderate $R^2$ of 0.62 and relatively high values for mean absolute error (MAE) of 2,118 kWh (7,227 kBtu) and RMSE 2,947 kWh (10,057 kBtu), indicating that while the model captures broad consumption trends, it struggles with finer-grained accuracy. The decline in performance highlights the loss of essential interactions and contextual variables due to feature reduction and feature quality, limiting the model's ability to generalize across the diverse housing stock represented in the dataset. Overall, these figures demonstrate that while the selected features offer interpretive clarity and reflect relevant domain factors, the complexity of residential energy use necessitates a more extensive and energy-use-related feature space for achieving high-fidelity forecasting.

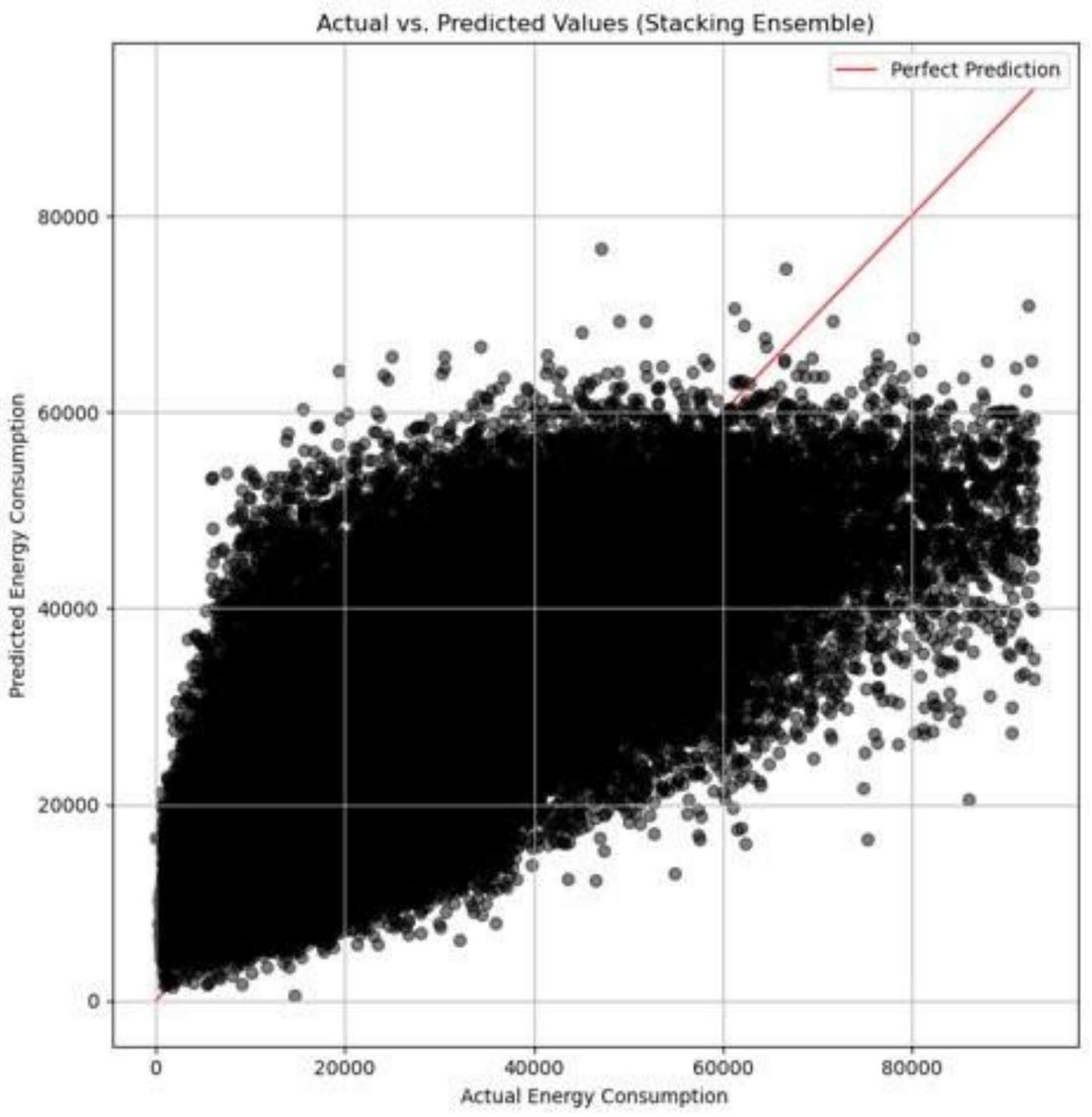

*Figure 15: Actual vs Predicted Scatter using the ten inputs from Table 5 for the ResStock dataset and the developed the CatBoost Machine Learning Model*

### 4.4. Data Slicing

Following the development of machine learning (ML) models for the RECS and ResStock datasets, that achieved less than 10% variance in forecasting residential energy consumption across the complete dataset, detailed in Sections 4.1 and 4.2 a more detailed analysis was conducted on the ResStock dataset to evaluate model performance on filtered subsets using the previously-developed best-performing CatBoost ML model along with a suite of other Machine Learning Models used for analysis in Sections 4.1 and 4.2. This secondary analysis sought to determine whether the observed trends and model effectiveness persisted when applied to more homogeneous segments of the data, which were constrained by key building and usage characteristics.

Specifically, the dataset was filtered according to the following parameters:

- Building Type: Single-Family Detached
- Climate Zone: 6A
- Primary Heating Fuel: Natural Gas Furnace
- Construction Vintage: 2000–2010

Applying these filters resulted in a subset of 1,194 data points. The same suite of ML models utilized in the broader analysis was deployed on this subset, with Lasso Regression emerging as the top-performing model following hyperparameter tuning. The top three models and their corresponding performance metrics are summarized in Table 7.

*Table 7: Performance metrics of the three best-performing models for the reduced inputs*

| Model Name | $R^2$ | MAE | RMSE |
|---|---|---|---|
| CatBoost | 0.829 | 5,293 | 6,809 |
| Lasso Regression | 0.849 | 5,042 | 6,397 |
| LightGBM | 0.825 | 5,326 | 6,886 |

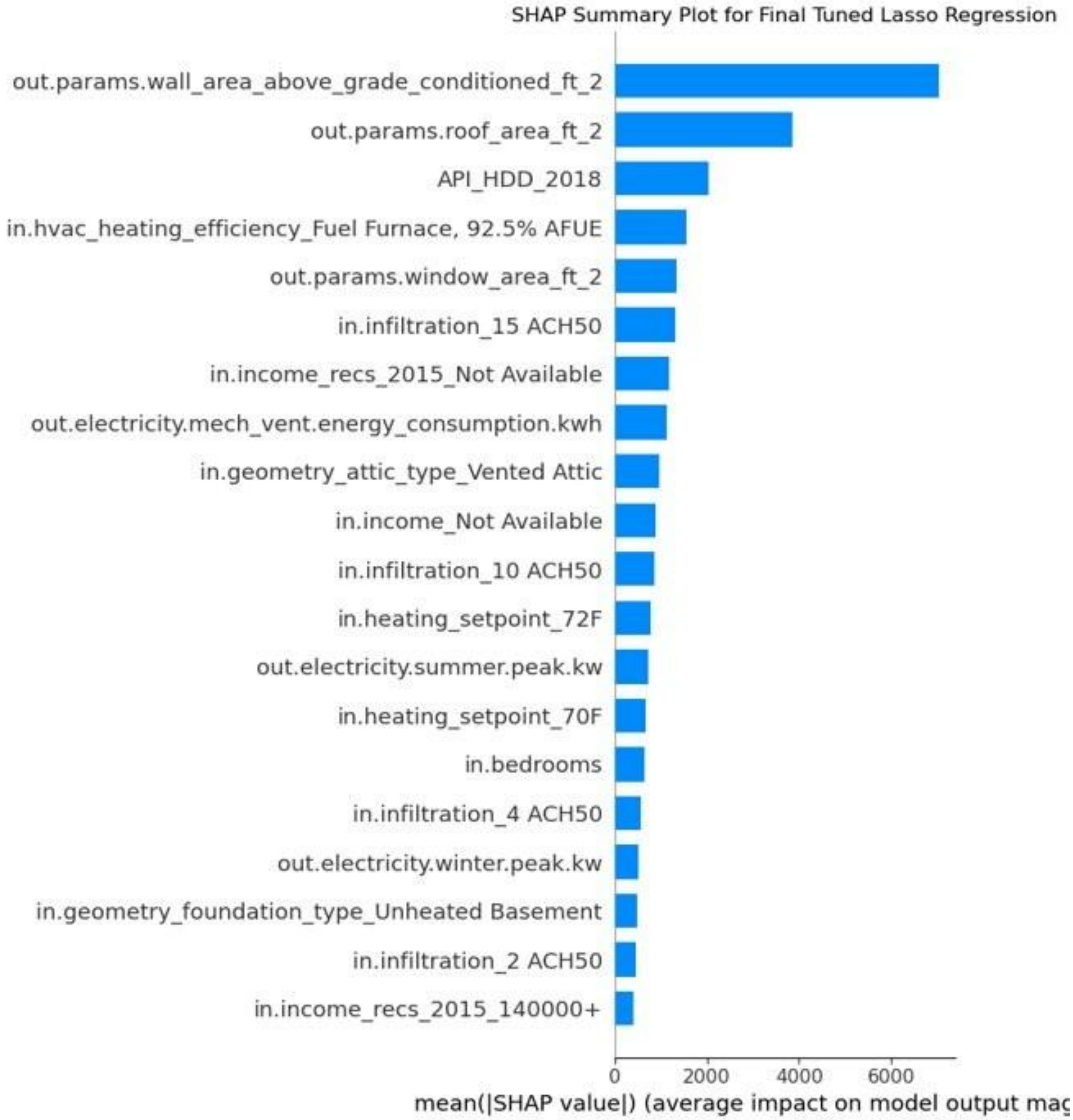

*Figure 16: SHAP plot showing the most correlated features for the filtered ResStock dataset using Lasso Regression*

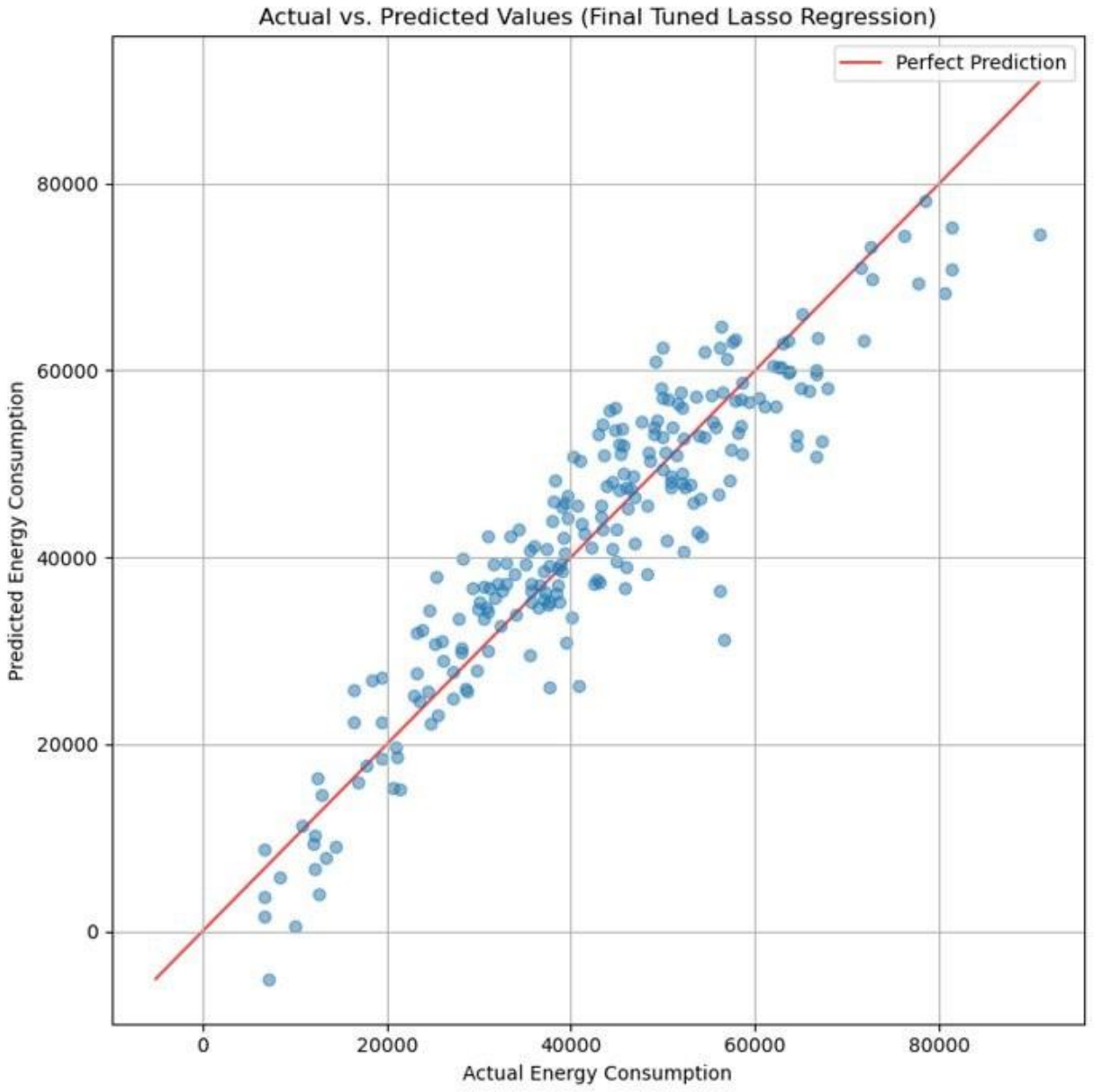

*Figure 17: Regression of the actual vs predicted total energy consumption values for the Lasso Regression Machine Learning model for the filtered ResStock dataset*

Conducting regression analysis on a nationwide scale inherently introduces variance due to the diversity within the building stock. In contrast to single-building calibration, where parameters are tightly controlled and detailed, a global dataset encompassing various climate zones, construction vintages, and fuel types establishes a natural ceiling for fit metrics, such as $R^2$, as illustrated by the variability in the Lasso regression analysis presented in Figure 17. While manually segmenting the data into homogeneous "bins". For instance, isolating Single-Family Detached homes in Climate Zone 6A would likely reduce variance and improve the coefficient of determination for those specific subsets, but such an approach undermines the model's generalizability. The "sweet spot" identified in this study emphasizes a cohesive architecture that can learn patterns across segments. By training on a comprehensive dataset, the Stacking Ensemble harnesses the extensive sample size to uncover macro-level dependencies, such as the non-linear effects of extreme weather events, which might remain statistically insignificant in smaller, segmented samples. Additionally, the Lasso regression machine learning model effectively automates the "binning" process by learning hierarchical feature interactions, thereby striking a balance between the fine granularity of manual segmentation and the scalability of a generalized national model.

To enhance ML-based analysis, SHAP was used to interpret feature importance, as illustrated in Figure 16, with regression performance shown in Figure 17. This disaggregated analysis serves a dual purpose: it not only validates the model's generalizability across subsets but also facilitates an examination of specific energy use patterns within more narrowly defined residential categories. For example, Figure 18 provides a visualization of the spread of energy consumption by floor area bin for the actual data in the ResStock dataset. Similarly, Figure 19 demonstrates the ML-predicted energy consumption is plotted by the floor area bins.

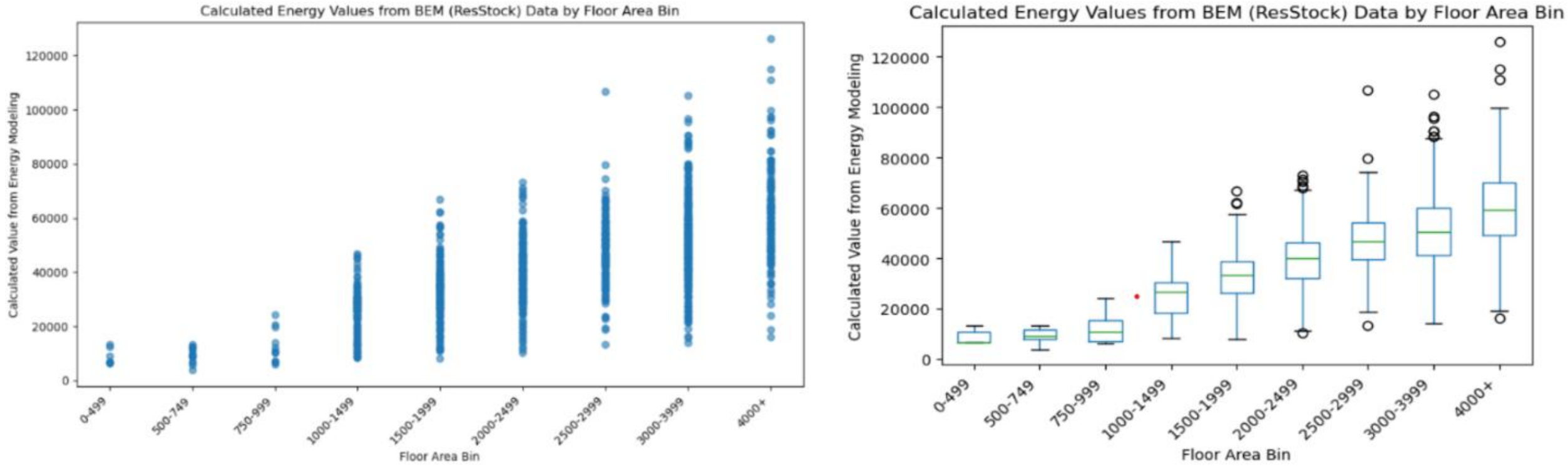

*Figure 18: ResStock modeled energy consumption as a function of floor area bins*

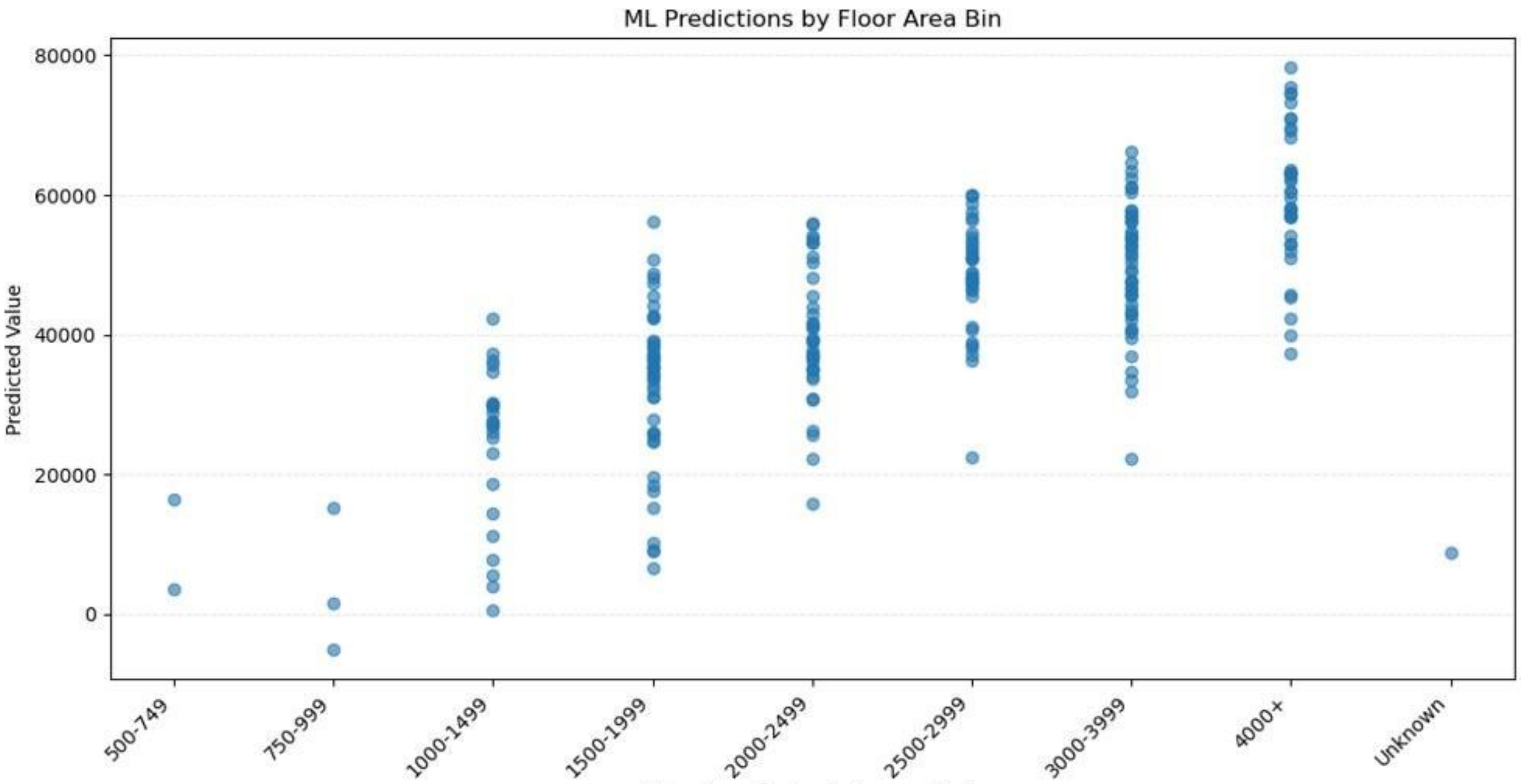

*Figure 19: Lasso Regression Machine Learning Model predicted energy consumption as a function of floor area bins*

A key focus of this exercise was to investigate how energy consumption varies with changes in floor area within the selected building and weather region. To this end, a linear regression was performed, modeling ResStock-simulated energy consumption as a function of floor area bins. The resulting regression equation (in the form y = mx + c) and corresponding $R^2$ values are presented in Figure 20, offering insights into the strength and nature of this relationship.

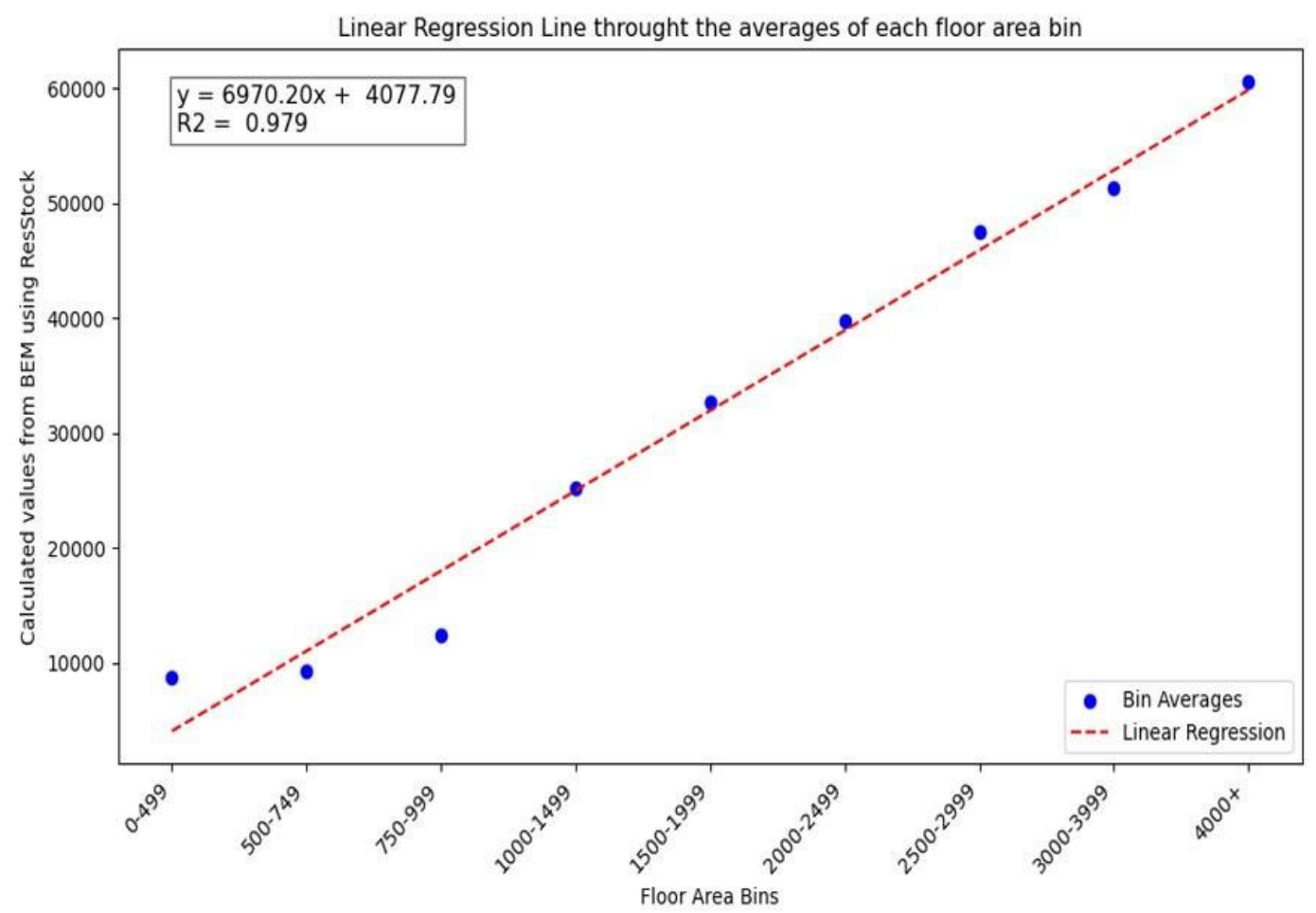

*Figure 20: Linear Regression through the average energy consumption per floor bin*

Figure 20 demonstrates the strong linear correlation between residential floor area and annual energy consumption, as derived from the ResStock Building Energy Model (BEM). The x-axis categorizes buildings by floor area bins, while the y-axis shows average energy consumption. The blue dots represent these averages, and the red dashed line indicates the best-fit linear regression, which has a high coefficient of determination ($R^2 = 0.979$). This suggests that floor area is a dominant first-order predictor of energy consumption. The regression slope (~6970) indicates that for each additional 1,000 square feet, annual energy use increases by about 6,970 kBtu (2,043 kWh), with a positive y-intercept (~4,078 kBtu (1,195 kWh)) that reflects base consumption from fixed end-uses. This strong correlation highlights the significance of floor area in energy modeling and its role in feature selection for predictive analysis.

# 5. Discussions

## 5.1. Key findings and interpretation

This study examined the application of advanced machine learning (ML) techniques to predict residential energy consumption, as well as space heating and cooling demands. We utilized two extensive datasets: the 2020 Residential Energy Consumption Survey (RECS) and the 2024.2 ResStock database, consisting of approximately 18,500 and 550,000 samples, respectively. Using thorough feature engineering, careful selection of relevant features, hyperparameter tuning, model stacking, and SHAP-based explanations, we developed models that outperform previously reported baselines on the same datasets. Additionally, we identified significant and physically plausible factors that influence energy performance. Ensemble tree models, particularly CatBoost, demonstrated exceptional performance, with $R^2$ values of approximately 0.905 for ResStock and 0.726 for RECS for models developed capable of predicting energy consumption. These results highlight the importance of using intuitive, directly correlated predictors when developing effective machine learning emulators for forecasting residential building energy performance. The CatBoost model proved to be particularly effective in predicting total energy consumption, as well as space heating and cooling, surpassing existing studies in the literature for both the RECS and ResStock datasets.

SHAP analyses reveal that important physical features such as heating and cooling loads, degree-day metrics, building geometry, home age, and primary fuel types influence predicted energy consumption in RECS. Heating-related features dominate total and heating forecasts, while cooling forecasts are driven by cooling loads, air-conditioning presence, and summer degree-days. This alignment with thermodynamic principles and existing research enhances confidence in the models' causal learning. Differences in predictive performance between the empirical and simulation-based datasets can be attributed to their distinct origins and differing levels of information granularity. ResStock's comprehensive, physics-based synthetic data yields the highest accuracy, while RECS, despite being nationally representative, presents greater noise due to the nature of the subjective user responses and is less complex in detail, reducing its predictive skill. Nevertheless, ResStock models are valuable for scenario analyses, although empirical calibration with RECS is crucial for real-world applications.

Limiting inputs to just ten types of homeowner responses about knowable building characteristics that don't necessarily conform to some of the features known and proven to have high correlation with energy use reduces accuracy, with $R^2$ values of approximately 0.61 for the RECS dataset and 0.62 for ResStock, compared to 0.73 and 0.90, respectively, using ML-determined features. This suggests that much of the predictive benefit comes from features not included in the restricted ten-variable input schedule, rather than from homeowner-accessibility alone. While limited-input models can capture basic outdoor temperature effects and home size, they miss crucial details that influence household-level variance. This finding is particularly relevant for applications seeking to minimize user effort, time, and cost, while recognizing that forecast accuracy depends not only on the number of inputs but also on the quality and value of those inputs. This limitation is particularly relevant in institutional settings where utility allowances are determined using simplified input structures, requiring models that can perform under constrained data availability. The study also examines how well the developed models can predict energy performance within a specific subset of data. For example, analyzing single-family detached homes in Climate Zone 6A with natural gas heating, built in the 2000s, results in improved model performance, with an $R^2$ value of approximately 0.85. This highlights how variability in building characteristics and occupant behavior can contribute to prediction errors in broader national models. Targeted models for specific groups can achieve high accuracy with fewer available predictors. This suggests a practical approach: using simpler models for cohort assignment, followed by more refined models tailored to those specific populations.

### 5.2. Limitations and future scope of work

As with any study, there exist limitations in the analysis. For example, both RECS and ResStock represent snapshots of buildings at a single point in time, which limits the models' ability to capture temporal variation or changes in energy use related to occupant behavior. High-performing models often depend on numerous hard-to-obtain variables, such as envelope information, infiltration measurements, mechanical equipment system attributes, and historical heating and cooling degree days (HDD/CDD) at zip code resolution. In practice, these variables may be unavailable or proprietary, leading to assumptions and/or inaccurate portrayal of information that might not be apt for different instances. Consequently, model performance declines when only homeowner-accessible inputs are utilized, complicating deployment for consumers or Public Housing Agencies (PHAs) and leading to potentially inaccurate forecasts and estimates.

The RECS relies on survey data, which is vulnerable to response bias, self-reporting errors, and billing-matching discrepancies. In contrast, ResStock is based on synthetic data; while it is physically detailed, it might underrepresent behavioral variance and anomalous equipment performance, thereby classifying it as "synthetic data". Models trained on these sources may inherit these biases, potentially leading to systematic errors for certain building archetypes or socio-demographic groups. Consequently, any decisions linked to these models, such as utility allowances, must include uncertainty quantification and bias checks. The observed decline in performance with reduced-feature models highlights the dependence of machine learning approaches on data richness and completeness, presenting challenges for real-world applications where detailed building and appliance information is frequently lacking or inaccurate, or where buildings undergo improvements that are not recorded. Furthermore, the RECS and ResStock datasets differ in population scope and stock construction; RECS mainly samples occupied primary residences only, whereas ResStock is a synthetic, weighted representation of the broader U.S. housing stock. These differences may limit generalizability to other housing markets.

Models incorporating historical HDD/CDD data and building characteristics often assume static occupant behaviors and weather-related baselines. However, changes in weather patterns and evolving behaviors, such as thermostat use and equipment type, necessitate periodic model recalibration to address potential drift. Key drivers of energy consumption, including thermostat schedules and occupancy patterns, are inherently behavioral and non-deterministic, resulting in an irreducible error floor for household-level forecasting due to the incomplete capture of these stochastic components by existing predictors. While SHAP analysis offers valuable insights for interpretability, it does not fully address the causal ambiguity inherent in correlational data. The integration of meticulous feature engineering, ensemble machine learning, and SHAP-based interpretability yields robust emulators of residential energy consumption for both survey- and physics-driven datasets. However, accuracy declines when limiting inputs to readily available homeowner data, underscoring a critical tension between model fidelity and data accessibility. Responsible implementation, particularly in public housing authority contexts, demands cohort-aware modeling, transparent uncertainty quantification, and continuous empirical validation.

Future research should prioritize the more accurate and complete characterization of building features by combining static survey-based data with temporal data streams, such as smart meter or sensor measurements, and, where available, more granular building and equipment descriptors. This integrated representation may better capture dynamic consumption patterns and improve prediction performance. Additionally, the development of data-efficient modeling strategies like transfer learning, federated learning, or physics-informed machine learning could support accurate predictions, even when faced with sparse input data. Incorporating uncertainty quantification is also essential for assessing model confidence and supporting subsequent decision-making.

# 6. Conclusions

This study evaluated residential energy estimation under two contrasting information conditions: information-rich modeling using detailed dataset-specific features and deployment-oriented modeling restricted to a predefined set of ten low-burden inputs. Using the 2020 Residential Energy Consumption Survey (RECS) and the 2024.2 ResStock dataset, machine-learning models were developed for annual total energy use, space heating, and space cooling. Under information-rich conditions, CatBoost provided the strongest overall performance among the evaluated algorithms, achieving an $R^2$ of 0.726 for total energy consumption in RECS and 0.905 in ResStock, with similarly strong performance for heating and cooling. The higher performance obtained with ResStock is consistent with its physically structured, high-resolution simulation inputs, whereas RECS reflects the greater variability and uncertainty associated with empirical household survey and billing data.

The principal finding of this study, however, concerns what happens when this information advantage is removed. When both datasets were restricted to the same ten practically obtainable variables, including degree-day indicators, housing type, number of bedrooms, heating and water-heating fuels, cooking fuel, and air-conditioning equipment, the difference between the two datasets largely disappeared. Total-energy estimation performance declined from $R^2 = 0.726$ to 0.61 for RECS and from $R^2 = 0.905$ to 0.62 for ResStock. The substantially larger reduction for ResStock indicates that much of the predictive advantage observed under information-rich conditions is associated with access to detailed physical, system, and operational descriptors that are unlikely to be available in many real-world deployment settings. Thus, performance obtained from comprehensive research datasets should not be assumed to persist when models must operate using only readily obtainable dwelling information.

The results therefore frame residential energy estimation as an accuracy–accessibility trade-off rather than solely an algorithm-selection problem. Detailed building information can support high-fidelity prediction, but collecting such information may require audits, extensive records, monitoring, or detailed simulation inputs. Conversely, low-burden models are substantially easier to deploy at scale but sacrifice household-level predictive accuracy. This distinction is particularly important for applications such as preliminary screening, portfolio-level assessment, benchmarking, and utility-allowance estimation, where detailed envelope, equipment-efficiency, infiltration, occupancy, and operational information may not be available for every dwelling. The purpose of the limited-input model is therefore not to replace detailed audits or calibrated simulations, but to quantify the level of predictive capability that can reasonably be retained when data collection itself is a practical constraint.

The cohort-specific analysis further demonstrated that part of the performance lost under limited information may be recovered when the model is applied to a more homogeneous residential population. For the selected ResStock cohort consisting of single-family detached homes in Climate Zone 6A, heated by natural gas and constructed between 2000 and 2010, the reduced-input analysis produced an $R^2$ of 0.849 with Lasso regression and 0.829 with CatBoost, compared with $R^2 = 0.62$ for the reduced-input national ResStock model.

Several limitations should be considered when interpreting the results. RECS and ResStock represent fundamentally different forms of evidence: RECS contains observations from actual households and therefore includes behavioral variability, reporting uncertainty, and billing-related noise, whereas ResStock provides a statistically representative but simulation-based characterization of the U.S. residential building stock. Models developed from either source may therefore inherit limitations associated with the underlying dataset. In addition, the present analyses are based primarily on cross-sectional annual characteristics and do not explicitly capture temporal changes in occupant behavior, equipment operation, renovations, or weather-dependent behavioral responses. These uncertainties are particularly important when limited-input models are used for consequential applications, where prediction uncertainty, population-specific bias, and empirical validation should accompany point estimates.

Overall, the study demonstrates that the information available at the time of deployment is a central determinant of achievable residential energy-estimation performance. The comparison between RECS and ResStock shows that substantial differences in predictive accuracy under information-rich conditions can narrow considerably when both models are subjected to the same practical information constraint. At the same time, the cohort analysis indicates that targeted modeling may recover meaningful predictive capability without requiring fully detailed building descriptions. Future work should therefore focus not only on improving modeling algorithms, but also on identifying efficient combinations of accessible variables, developing systematic cohort-selection strategies, quantifying prediction uncertainty, and validating reduced-input models against independent empirical datasets. Such efforts could support residential energy-estimation tools that explicitly balance predictive fidelity against the cost and feasibility of obtaining the information required for real-world deployment.